\documentclass[acmtog]{acmart}

\acmSubmissionID{472}

\usepackage{booktabs} 

\usepackage{dsfont} 

\setcitestyle{square}

\usepackage[ruled]{algorithm2e} 

\SetAlFnt{\small}
\SetAlCapFnt{\small}
\SetAlCapNameFnt{\small}
\SetAlCapHSkip{0pt}
\IncMargin{-\parindent}

\acmJournal{TOG}

\acmJournal{TOG}
\acmYear{2020}\acmVolume{39}\acmNumber{4}\acmArticle{62}\acmMonth{7} \acmDOI{10.1145/3386569.3392462}

\newcommand{\etal}{et al.}
\usepackage{color}
\usepackage{soul}
\usepackage{breqn}
\usepackage{booktabs}
\usepackage{tabularx}
\usepackage{rotating}

\definecolor{purple}{rgb}{0.65,0,0.65}
\definecolor{blue}{rgb}{0, 0.2, 0.8}
\definecolor{orange}{rgb}{0.6, 0.6, 0}
\definecolor{red}{rgb}{0.8, 0.2, 0.2}
\definecolor{magenta}{rgb}{0.5, 0.0, 1.0}
\definecolor{black}{rgb}{0.0, 0.0, 0.0}
\definecolor{cyan}{rgb}{0, 0.65, 0.65}
\definecolor{olive}{rgb}{0.2, 0.6, 0.5}
\usepackage{xcolor}

\newif\ifdraft
\drafttrue

\ifdraft
\newcommand{\dlc}[1]{{\color{magenta}\textbf{DL:} #1}}
\newcommand{\dcc}[1]{{\color{red}\textbf{DC:} #1}}
\newcommand{\kac}[1]{{\color{orange}\textbf{KA:} #1}}
\newcommand{\plc}[1]{{\color{blue}\textbf{PL:} #1}}

\else
\newcommand{\dlc}[1]{}
\newcommand{\dcc}[1]{}
\newcommand{\kac}[1]{}
\newcommand{\plc}[1]{}
\newcommand{\osc}[1]{}

\fi

\newcommand{\bbm}{{\bf M}}
\newcommand{\bbp}{{\bf P}}
\newcommand{\bbq}{{\bf Q}}
\newcommand{\bbr}{{\bf R}}
\newcommand{\bbs}{{\bf S}}

\newcommand{\bb}{{\bf b}}

\newcommand{\bbw}{{\bf W}}

\newcommand{\FK}{\text{FK}}
\newcommand{\Loss}{\mathcal{L}}
\newcommand{\mm}{\mathcal{M}}
\newcommand{\mmn}{\mathcal{N}}
\newcommand{\mms}{\mathcal{S}}

\newcommand{\RR}{\mathds{R}}
\newcommand{\bbe}{\mathbb{E}}
\def \figures {./}

\begin{document}

\title{Skeleton-Aware Networks for Deep Motion Retargeting}

\author{Kfir Aberman}
\affiliation{\institution{AICFVE, Bejing Film Academy \& Tel-Aviv University}}
\authornote{equal contribution}

\author{Peizhuo Li}
\affiliation{\institution{CFCS, Peking University \& AICFVE, Beijing Film Academy}}
\authornotemark[1]

\author{Dani Lischinski}
\affiliation{%
    \institution{The Hebrew University of Jerusalem \& AICFVE, Beijing Film Academy}
}
\author{Olga Sorkine-Hornung}
\affiliation{\institution{ETH Zurich \& AICFVE, Beijing Film Academy}}

\author{Daniel Cohen-Or}
\affiliation{\institution{Tel-Aviv University \& AICFVE, Beijing Film Academy}}

\author{Baoquan Chen}
\affiliation{\institution{CFCS, Peking University \& AICFVE, Beijing Film Academy}}
\authornote{corresponding author}

\renewcommand\shortauthors{Aberman, K. et al.}

\authorsaddresses{%
Authors' addresses: Kfir Aberman, kfiraberman@gmail.com; Peizhuo Li, peizhuo@pku.edu.cn; Dani Lischinski, danix3d@gmail.com; Olga Sorkine-Hornung, sorkine@inf.ethz.ch; Daniel Cohen-Or, cohenor@gmail.com; Baoquan Chen, \mbox{baoquan@pku.edu.cn}}

\begin{abstract}
	
We introduce a novel deep learning framework for data-driven motion retargeting between skeletons, which may have different structure, yet corresponding to homeomorphic graphs.
Importantly, our approach learns how to retarget without requiring any explicit pairing between the motions in the training set.

We leverage the fact that different homeomorphic skeletons may be reduced to a common \emph{primal skeleton} by a sequence of edge merging operations, which we refer to as \emph{skeletal pooling}.
Thus, our main technical contribution is the introduction of novel differentiable convolution, pooling, and unpooling operators.
These operators are \emph{skeleton-aware}, meaning that they explicitly account for the skeleton's hierarchical structure and joint adjacency, and together they serve to transform the original motion into a collection of deep temporal features associated with the joints of the primal skeleton. 
In other words, our operators form the building blocks of a new deep motion processing framework that embeds the motion into a common latent space, shared by a collection of homeomorphic skeletons. 
Thus, retargeting can be achieved simply by encoding to, and decoding from this latent space.

Our experiments show the effectiveness of our framework for motion retargeting, as well as motion processing in general, compared to existing approaches.
Our approach is also quantitatively evaluated on a synthetic dataset that contains pairs of motions applied to different skeletons.
To the best of our knowledge, our method is the first to perform retargeting between skeletons with differently sampled kinematic chains, without any paired examples.
	
\end{abstract}

%

%
%



\begin{teaserfigure}
\centering
\includegraphics[width=\linewidth]{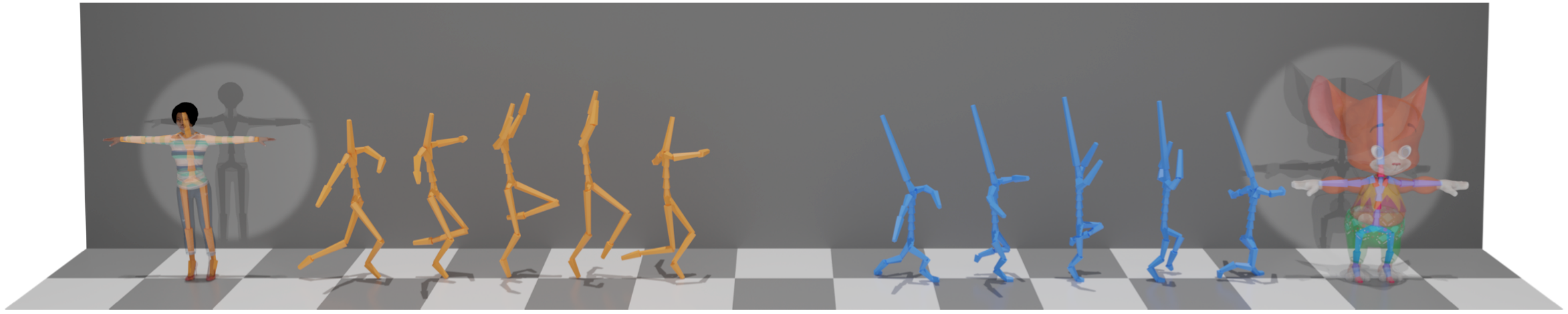} 
\caption{Unpaired, cross-structural, motion retargeting. An input motion sequence (orange skeletons) is retargeted to a target skeleton (rightmost, blue), which has different body proportions, as well as a different number of bones (marked in red).}
\label{fig:teaser}
\end{teaserfigure}

\maketitle

\section{Introduction}
\label{sec:intro}
Capturing the motion of humans is a fundamental task in motion analysis, computer animation, and human-computer interaction.
Motion capture (MoCap) systems typically require the performer to wear a set of markers, whose positions are sampled by magnetic or optical sensors, yielding a temporal sequence of 3D skeleton poses.
Since different MoCap setups involve different marker configurations and make use of different software, the captured skeletons may differ in their structure and number of joints, in addition to differences in bone lengths and proportions, corresponding to different captured individuals.
Thus, \emph{motion retargeting} is necessary, not only for transferring captured motion from one articulated character to another, within the same MoCap setup, but also across different setups. The latter is also essential for using data from multiple different motion datasets in order to train \emph{universal}, setup-agnostic, data-driven models, for various motion processing tasks. 

Deep neural networks, which revolutionized the state-of-the-art for many computer vision tasks, leverage the regular grid representation of images and video, which is well suited for convolution and pooling operations. Unlike images, skeletons of different characters exhibit irregular connectivity. Furthermore, the structure of a skeleton is typically hierarchical. These differences suggest that the existing operators commonly used in CNNs might not be the best choice for analysis and synthesis of articulated motion.

In this paper, we introduce a new motion processing framework consisting of a representation for motion of articulated skeletons, designed for deep learning, and several differentiable operators, including convolution, pooling and unpooling, that operate on this representation. The operators are \emph{skeleton-aware}, which means that they explicitly account for the skeleton structure (hierarchy and joint adjacency). These operators constitute the building blocks of a new deep framework, where the shallower layers learn local, low-level, correlations between joint rotations, and the deeper layers learn higher-level correlation between body parts. 

The proposed motion processing framework can be useful for various motion analysis and synthesis learning based tasks. In this work, we focus on the task of motion retargeting between skeletons that have the same set of end-effectors, but might differ in the number of joints along the kinematic chains from the root to these end effectors. Such skeletons may be represented by homeomorphic (topologically equivalent) graphs.

Although motion retargeting is a long-standing problem, current approaches cannot automatically perform retargeting between skeletons that differ in their structure or the number of joints \cite{villegas2018neural}.
In this scenario, correspondences between the different skeletons should be manually specified, often resulting in unavoidable retargeting errors. Animators must then manually correct such errors by manipulating key frames, which is a highly tedious process.

We treat the retargeting problem as a multimodal translation between unpaired domains, where each domain contains a collection of motions performed by skeletons with different proportions and bone lengths that share a specific skeletal structure (same set of kinematic chains). Thus, all of the motions in a given domain may be represented using the same graph. Different domains contain skeletons with different structure, i.e., represented by different graphs; however, the graphs are assumed to be homeomorphic.

Previous work has demonstrated that multimodal unpaired image translation tasks may be carried out effectively using a shared latent space \cite{huang2018multimodal, gonzalez2018image}. In these works, same-sized images from different domains are embedded in a shared space that represents, for example, the content of the image, disentangled from its style. On images, such an embedding is straightforward to carry out using standard convolution and pooling operators; however, this is not the case for skeletons with different structures. In this work, we utilize our skeleton-aware motion processing framework, in particular \emph{skeletal pooling}, to embed motions performed by different skeletons into a shared latent domain space.

Our key idea exploits the fact that different, yet homeomorphic, skeletons may be reduced to a common \emph{primal skeleton}, which may be viewed as the common ancestor of all the different skeletons in the training data, by merging pairs of adjacent edges/armatures. Through interleaving of skeletal convolution and pooling layers, the shared latent space consists of a collection of deep temporal features, associated with the joints of the primal skeleton. The latent space is jointly learned by an encoder-decoder pair for each skeletal structure (domain). 

In addition, we exploit our deep motion representation to disentangle the motion properties from the shape properties of the skeleton, which allows us to perform motion retargeting using a simple algorithm in our deep feature space.
Similarly to the raw, low-level, representation of motion, which consists of static (joint offsets) and dynamic (joint rotations) components, our deep motion features are also split into static and dynamic parts. However, in the raw input motion, the two components are strongly coupled: a specific sequence of joint rotations is bound to a specific bone length and skeleton structure. In contrast, our encoders learn to decouple them: the dynamic part of the latent code becomes skeleton-agnostic, while the static one corresponds to the common primal skeleton. This property of the latent space makes it possible to retarget motion from skeleton $A$ to skeleton $B$ simply by feeding the latent code produced by the encoder of $A$ into the decoder of $B$.

In summary, our two main contributions in this work are:
\begin{enumerate}
	\item A new motion processing framework that consists of a deep motion representation and differentiable skeleton-aware convolution, pooling, and unpooling operators.
    \item A new architecture for unpaired motion retargeting between topologically equivalent skeletons that may have different number of joints.
\end{enumerate}

In addition to presenting the first automatic method for retargeting between skeletons with different structure, without any paired examples, we also demonstrate the effectiveness of our new deep motion processing framework for motion denoising (Section \ref{sec:denoising}).
We evaluate our motion retargeting approach and compare it to existing approaches in Section \ref{sec:experiments}. A synthetic dataset, which contains pairs of motions applied to different skeletons, is used to perform a quantitative evaluation of our method. 
\section{Related Work}
\label{sec:relatedwork}

\subsection{Motion Retargeting}

In their pioneering work, Gleicher \etal~\shortcite{gleicher1998retargetting} tackled character motion retargeting by formulating a spacetime optimization problem with kinematic constraints, and solving it for the entire motion sequence. Lee and Shin \shortcite{lee1999hierarchical} explored a different approach, which first applies inverse kinematics (IK) at each frame to satisfy constraints, and then smooths the resulting motion by fitting multilevel B-spline curves. The online retargeting method of Choi and Ko \shortcite{choi2000online} performs IK at each frame and computes the change in joint angles corresponding to the change in end-effector positions, while ensuring that the high-frequency details of the original motion are well preserved. In their physically-based motion retargeting filter, Tak and Ko \shortcite{tak2005physically} make use of dynamics constraints to achieve physically plausible motions. Feng~\etal~\shortcite{feng2012automating} proposed heuristics that can map arbitrary joints to canonical ones, and describe an algorithm that enables to instill a set of behaviors onto an arbitrary humanoid skeleton.

Classical motion retargeting approaches, such as the one mentioned above, rely on optimization with hand-crafted kinematic constraints for particular motions, and involve simplifying assumptions. Increased availability of captured motion data has made data-driven approaches more attractive.
Delhaisse \etal~\shortcite{delhaisse2017transfer} describe a method for transferring the learnt latent representation of motions from one robot to another.
Jang \etal~\shortcite{jang2018variational} train a deep autoencoder, whose latent space is optimized to create the desired change in bone lengths.
These methods require paired training data.

Inspired by methods for unpaired image-to-image translation \cite{zhu2017unpaired},
Villegas \etal~\shortcite{villegas2018neural} propose a recurrent neural network architecture with a Forward Kinematics layer and cycle consistency based adversarial training objective for motion retargeting using unpaired training data.
Lim \etal~\shortcite{lim2019PMnetLO} report better results for the unpaired setting by learning to disentangle pose and movement.
Aberman~\etal~\shortcite{aberman2019learning} disentangle 2D human motion extracted from video into character-agnostic motion, view angle and skeleton, enabling retargeting of 2D motion while bypassing 3D reconstruction.
All of these data-driven methods assume that the source and target articulated structures are the same.

Several works have explored retargeting human motion data to non-humanoid characters \cite{yamane2010animatingNC,seol2013creatureFO}.
In this scenario, the source and target skeletons may differ greatly from each other, however, the above approaches require captured motions of a human subject acting in the style of the target character.
It is also necessary to select a few key poses from the captured motion sequence and match them to corresponding character poses \cite{yamane2010animatingNC}, or pair together corresponding motions \cite{seol2013creatureFO}, in order to learn the mapping.
Celikcan \etal~\shortcite{celikcan2015exampleBasedRO} retarget human motion to arbitrary mesh models, rather than to skeletons. Similarly to the aforementioned approaches, learning such a mapping requires a number of pose-to-pose correspondences.

Abdul-Massih \etal~\shortcite{abdulmassih2017motionSR} argue that the motion style of a character may be represented by the motions of groups of body parts (GBPs). Thus, motion style retargeting may be done across skeleton structures by establishing a correspondence between GBPs, followed by constrained optimization to preserve the original motion. This requires defining the GBPs and the correspondence between them for each pair of characters.

Another loosely related problem is that of mesh deformation transfer. Earlier works, e.g.,
\cite{sumner2004deformation,baran2009semanticDT} require multiple correspondences between the meshes. The recent method of Gao \etal~\shortcite{gao2018automaticUS} uses unpaired datasets to train a VAE-CycleGAN that learns a mapping between shape spaces. We also leverage adversarial training to avoid the need for paired data or correspondences, but the setting of our work is somewhat different, since we deal with temporal sequences of hierarchical articulated structures (the skeletons), rather than meshes.

\subsection{Neural Motion Processing}

Holden \etal~\shortcite{holden2015learning,holden2016deep} were among the first to apply CNNs to 3D character animation. Motion is represented as a temporal sequence of 3D joint positions, and convolution kernels are local along the temporal dimension, but global in the joints dimension (the support includes all of the skeleton's joints). Thus, joint connectivity and the hierarchical structure of the skeleton are ignored. 

Furthermore, representing motion using 3D joint positions doesn't fully describe motion, and requires IK to extract an animation. Pavllo \etal~\shortcite{pavllo2019modelingHM} proposed QuaterNet, which processes joint rotations (quaternions), but performs forward kinematics on a skeleton to penalize joint positions, rather then angle errors. However, convolutional features extracted from joint rotations alone cannot fully capture 3D motion, since the same set of joint quaternions results in different poses when applied to different skeletons.

Since the skeleton of an articulated character may be represented as a graph, one might consider using graph convolutional networks (GCNs) to process motion data. In such networks, convolution filters are applied directly on graph nodes and their neighbors (e.g., \cite{bruna2013spectral,Niepert:2016}). Yan \etal~\shortcite{yan2018spatial} propose a spatial-temporal GCN (ST-GCN) for skeleton-based action recognition, where spatial convolution filters are applied to the 1-neighbors of each node in the skeleton, and temporal convolution is applied on successive positions of each joint in time. In this approach, no pooling takes place in the spatial dimension, but only in the temporal one. Furthermore, the set of 1-neighbors of each node is split into several sets, according to a hand-crafted strategy, in order to assign each set with a learnable convolution weight. Ci \etal~\shortcite{ci2019optimizingNS} apply a Locally Connected Network (LCN), which generalizes both fully connected networks and GCNs, to tackle 3D pose estimation. This work does not address processing of 3D motion.

Another option, which is not convolution-based, is to perform learning on spatio-temporal graphs using deep RNNs \cite{jain2015structuralRNNDL}. Wang \etal~\shortcite{wang2019spatiotemporal} propose a spatio-temporal RNN, where the skeletons are encoded into a latent space using a hierarchical neural network on the skeleton graph structure. The hierarchy is hand-crafted, while the weights of the fully-connected layers that merge different nodes together are learned. Thus, neither skeleton convolution, nor skeleton pooling takes place. The goals of this approach are also different from ours (prediction of motion, rather than its retargeting).

\section{Overview}
\label{sec:overview}

\begin{figure}
	\centering
	\includegraphics[width=\linewidth]{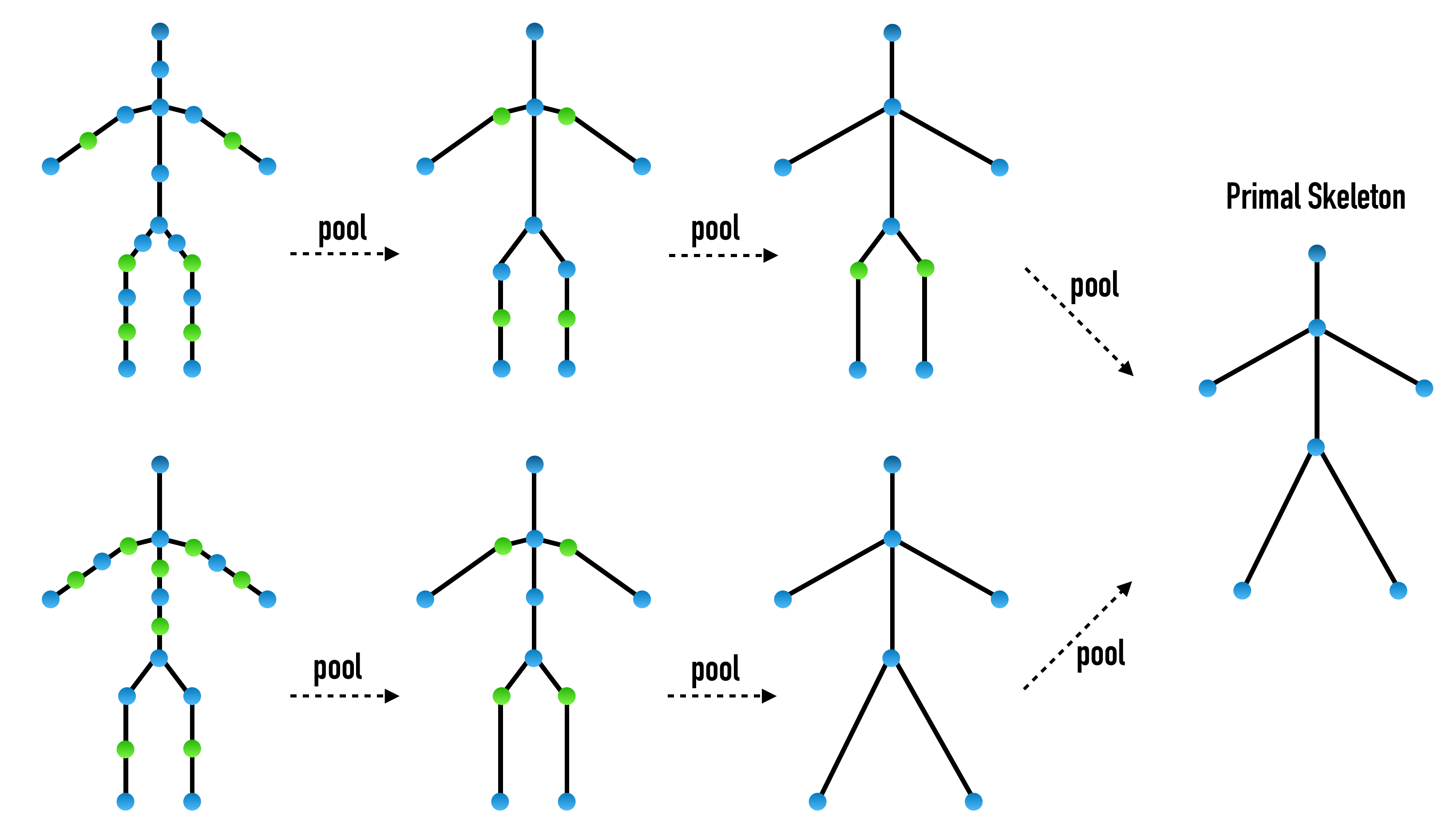} 
	\caption{Pooling to the primal skeleton. Our pooling operator removes nodes of degree two (green points) and merges their adjacent edges. After a few pooling steps the skeletal structures of different, topologically equivalent, skeletons are reduced into a common primal skeleton.}
	\label{fig:pool_to_primal}
\end{figure}

Our goal is to cope with the task of motion retargeting between skeletons that may have different structure, but are topologically equivalent.
The key idea is to exploit the fact that topologically equivalent skeletons may be represented by homeomorphic graphs. A pair of such graphs may be reduced into a common minimal graph by eliminating nodes of degree 2 along linear branches, as illustrated in Figure~\ref{fig:pool_to_primal}.
We refer to the skeleton represented by the reduced common graph as the \emph{primal skeleton}.

\begin{figure}
\centering
	\includegraphics[width=\linewidth]{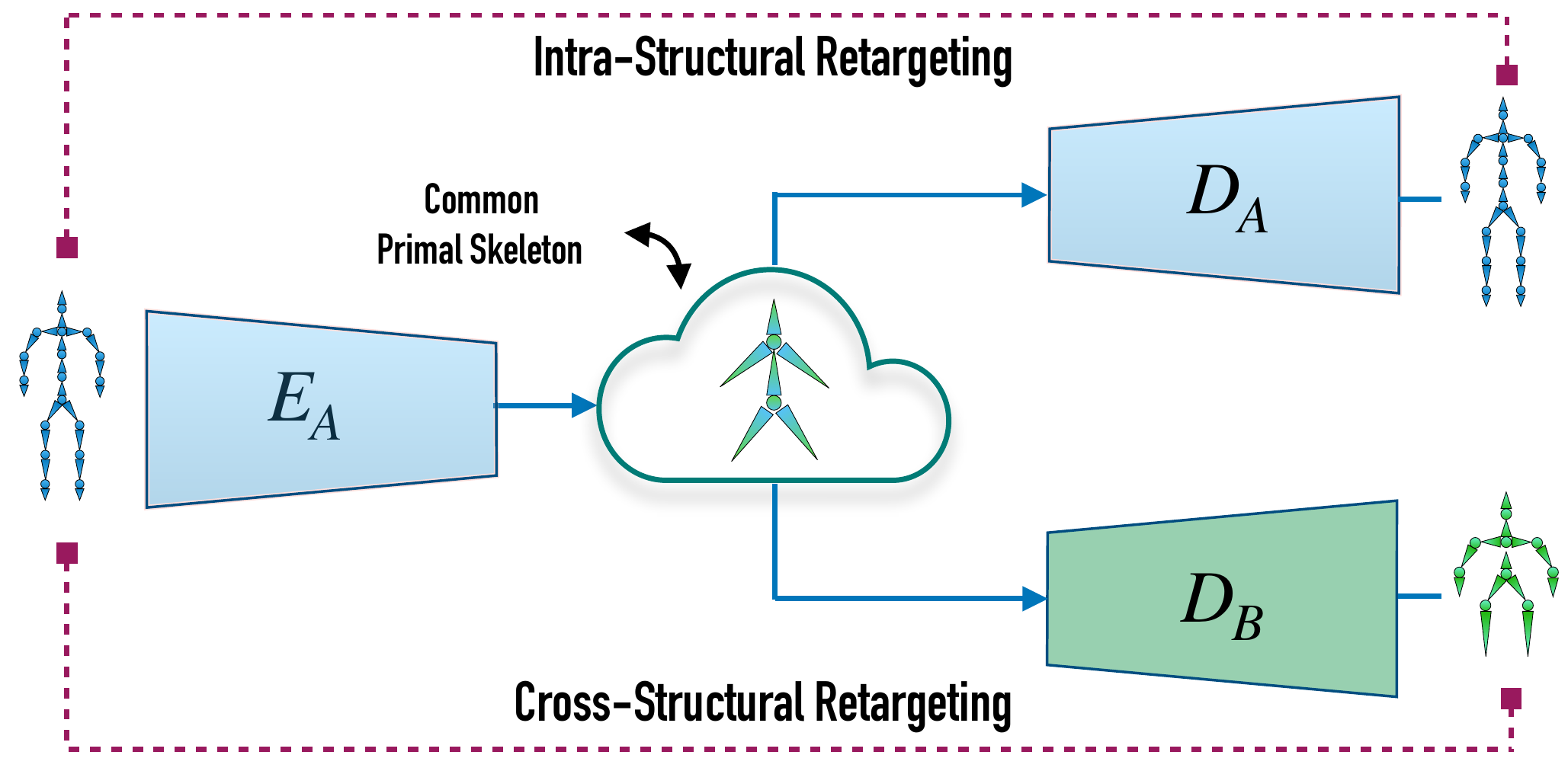} 
	\caption{Unpaired cross-structural motion retargeting. Our architecture encodes motions of different homeomorphic skeletons into a shared deep latent space, corresponding to a common primal skeleton. This representation can then be decoded into motions performed by skeletons within the same domain (intra-structural retargeting) or from another homeomorphic domain (cross-structural retargeting).
	}
	\label{fig:shared_latent}
\end{figure}

This observation suggests encoding motions performed by different homeomorphic skeletons into a deep representation that is independent of the original skeletal structure or bone proportions.
The resulting latent space is thus common to motions performed by skeletons with different structure, and we use it to learn data-driven motion retargeting, without requiring any paired training data. The retargeting process, depicted in Figure ~\ref{fig:shared_latent}, uses an encoder $E_A$ trained on a domain of motions, performed by skeletons with the same structure, to encode the source motion into the common latent space.
From this space, the latent representation may be decoded into a motion performed by a target skeleton. The target skeleton might have the same structure, but different bone lengths, in which case the decoding is done by a decoder $D_A$, trained on the same domain.
However, using a decoder $D_B$ trained on a different domain, the motion may also be retargeted across different skeletal structures.

In order to implement the approach outlined above, we introduce a new deep motion processing framework, consisting of two novel components: \\
(1) \textbf{Deep Motion Representation.} We represent a motion sequence as a temporal set of \emph{armatures} that constitute a graph, where each armature is represented by a dynamic, time-dependent, feature vector (usually referred to as joint rotation) as well as a static, time-independent one (usually referred to as offset), as depicted in Figure~\ref{fig:motion_representation}. The static-dynamic structure is a common low-level representation in character animation, which our framework preserves along the processing chain. Specifically, we use two branches (static and dynamic) that convert the low-level information into a deep, static-dynamic representation of motion features.\\
(2) \textbf{Deep Skeletal Operators.} We define new differential operators that can be applied to animated skeletons. The operators are \emph{skeleton-aware}, namely, the skeletal structure (hierarchy and joint adjacency) is considered by the operators. Concatenating these operators into an optimizable neural network enables the learning of deep temporal features that represent low-level, local joint correlations in shallow layers and high-level, global body part correlations in deeper layers.

Our motion processing framework, including the new representation and the skeleton-aware operators, is described in Section~\ref{sec:skeleton_aware}, while the architecture and loss functions that enable data-driven cross-structural motion retargeting are described in Section~\ref{sec:retargeting}.

\section{Skeleton-Aware Deep Motion Processing}
\label{sec:skeleton_aware}
Below we describe our motion processing framework, which consists of our motion representation, as well as our new skeletal convolution, pooling, and unpooling operators. 

\subsection{Motion Representation}
\label{subsec:motionrep}
\begin{figure}
	\centering
	\includegraphics[width=\linewidth]{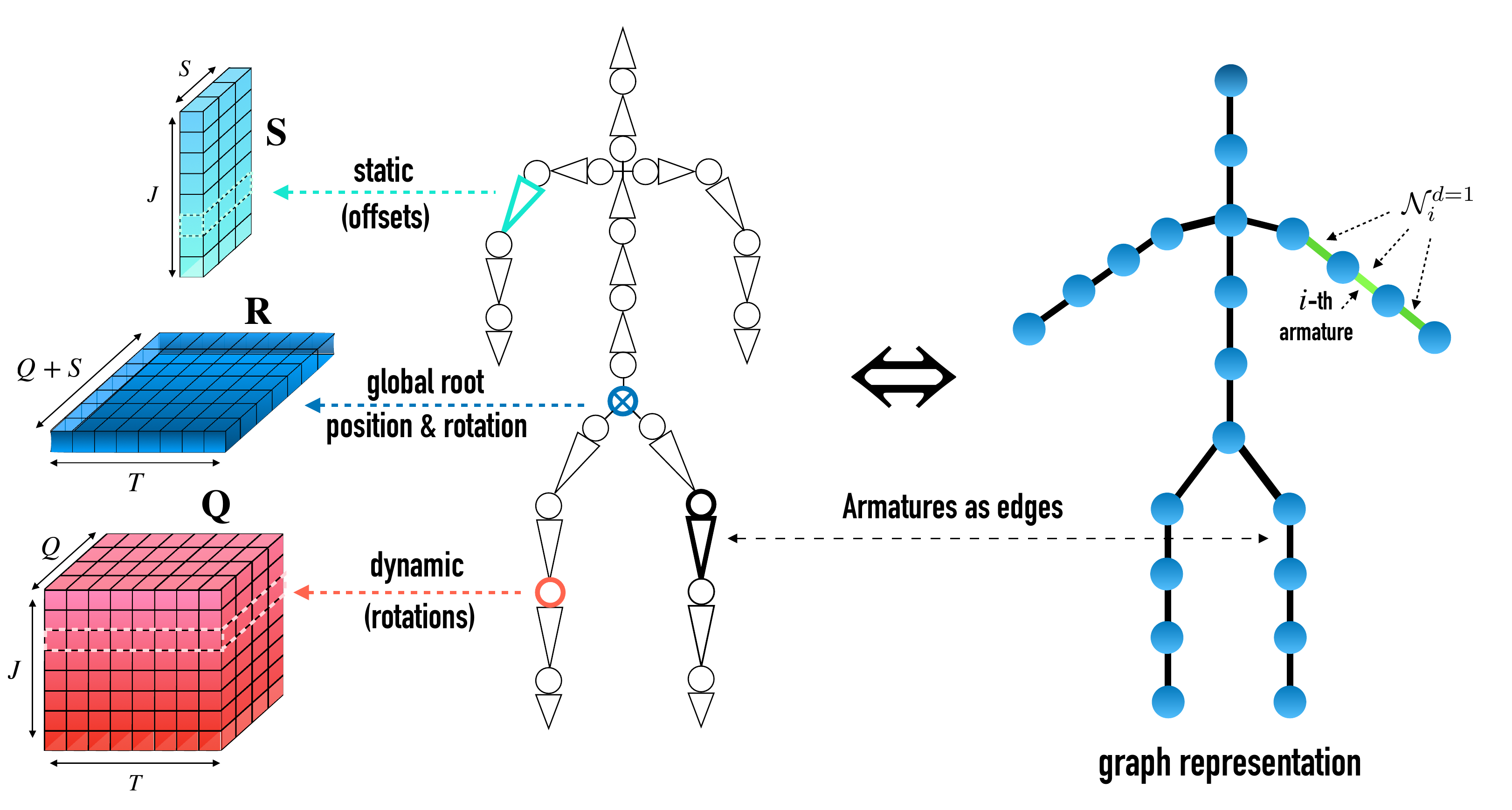} 
	\caption{We represent a motion clip (with temporal length $T$) of an articulated character by a set of $J$ armatures (middle), each described by a static $\bbs$ (cyan) and a dynamic $\bbq$ (red) feature vector. Additionally, we  store a sequence of global root positions and orientations $\bbr$ (blue). The skeleton is represented as a graph (right), where each edge $i$ corresponds to an armature with an adjacency list $\mmn_i^d$, defined based on the kinematic chain.}
	\label{fig:motion_representation}
\end{figure}

Our motion representation for articulated characters is illustrated in Figure~\ref{fig:motion_representation}. 
A motion sequence of length $T$ is described by a static component $\bbs\in\RR^{J \times S}$ and a dynamic one $\bbq\in\RR^{T\times J \times Q}$, where $J$ is the number of armatures, and $S$ and $Q$ are the dimensions of the static and the dynamic features, respectively (typically, $S=3$ and $Q=4$). 

The static component $\bbs$ consists of a set of offsets (3D vectors), which describe the skeleton in some arbitrary initial pose, while the dynamic component $\bbq$ specifies the temporal sequences of rotations of each joint (relative to its parent's coordinate frame in the kinematic chain), represented by unit quaternions.
The root joint $\bbr\in\RR^{T \times (S+Q)}$ is represented separately from the $J$ armatures (its children), as a sequence of global translations and rotations (orientations). 

The skeleton structure is represented by a tree graph whose nodes correspond to joints and end-effectors, while the edges correspond to armatures, as illustrated in Figure~\ref{fig:motion_representation}. Thus, for a skeleton with $J$ armatures, the graph has $J+1$ nodes. Connectivity is determined by the kinematic chains (the paths from the root joint to the end-effectors) and expressed by adjacency lists ${\bf \mmn}^d = \{\mmn_1^d, \mmn_2^d, \ldots, \mmn_J^d \}$, where $\mmn_i^d$ denotes the edges whose distance in the tree is equal or less than $d$ from the $i$-th edge (see Figure~\ref{fig:motion_representation}). 

\subsection{Skeletal Convolution}
\label{subsec:skeletalconv}

\begin{figure}
	\centering
	\includegraphics[width=\linewidth]{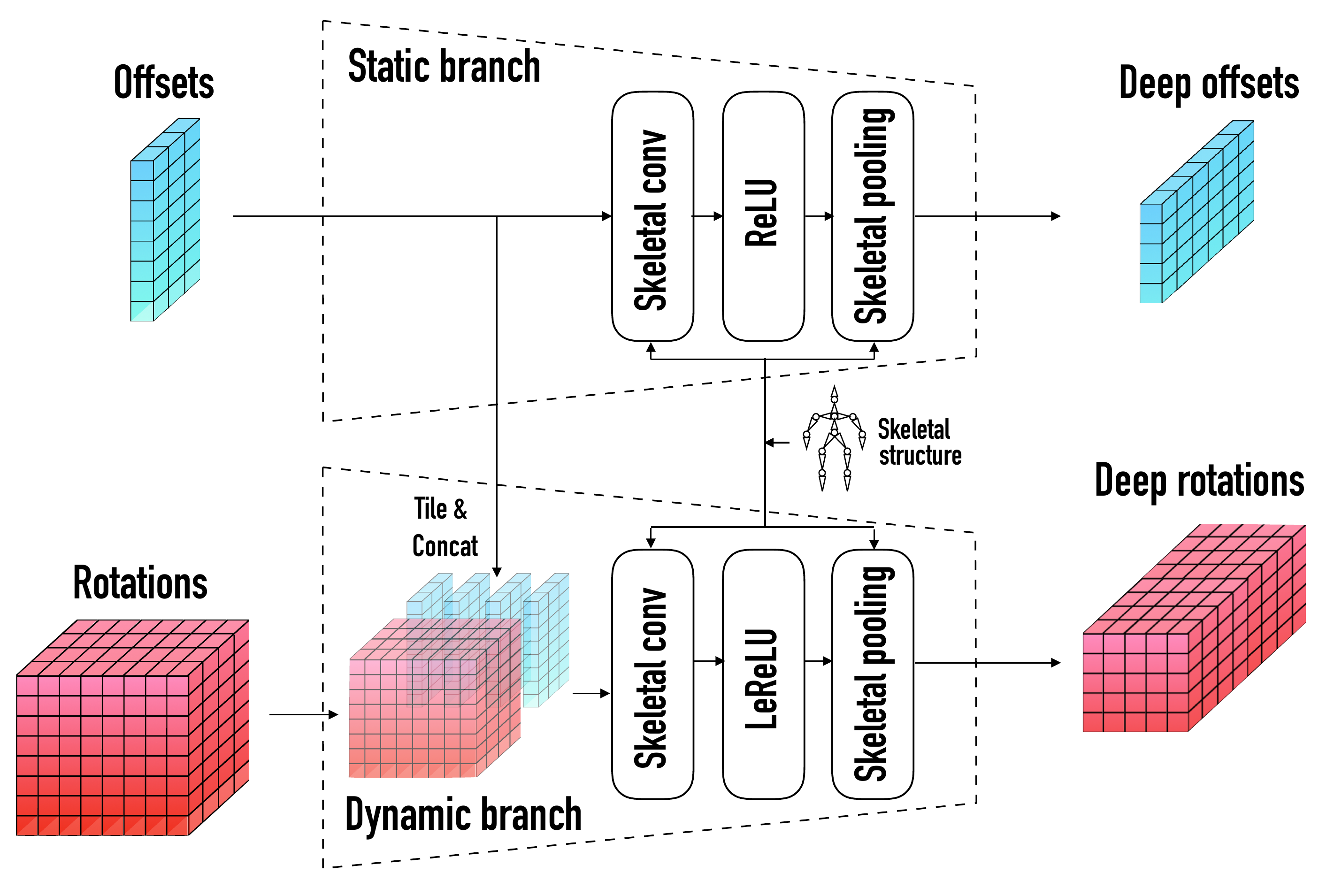} 
	\caption{Our framework performs skeletal convolution, activation, and skeletal pooling using blocks consisting of two parallel branches, dynamic and static. The dynamic branch takes tiled and concatenated static features as part of the input to its skeleto-temporal convolution layer. The static branch operates only on the static features.
	\label{fig:skeletal_block}
	}
\end{figure}

We process motion in two parallel branches: a dynamic branch, which yields time-dependent deep features, and a static branch, which produces static deep features. Both branches share the property that their convolution kernels consider the skeleton structure to compute local features across armatures, as illustrated in Figure~\ref{fig:skeletal_block}.

\begin{figure}
	\centering
	\includegraphics[width=\linewidth]{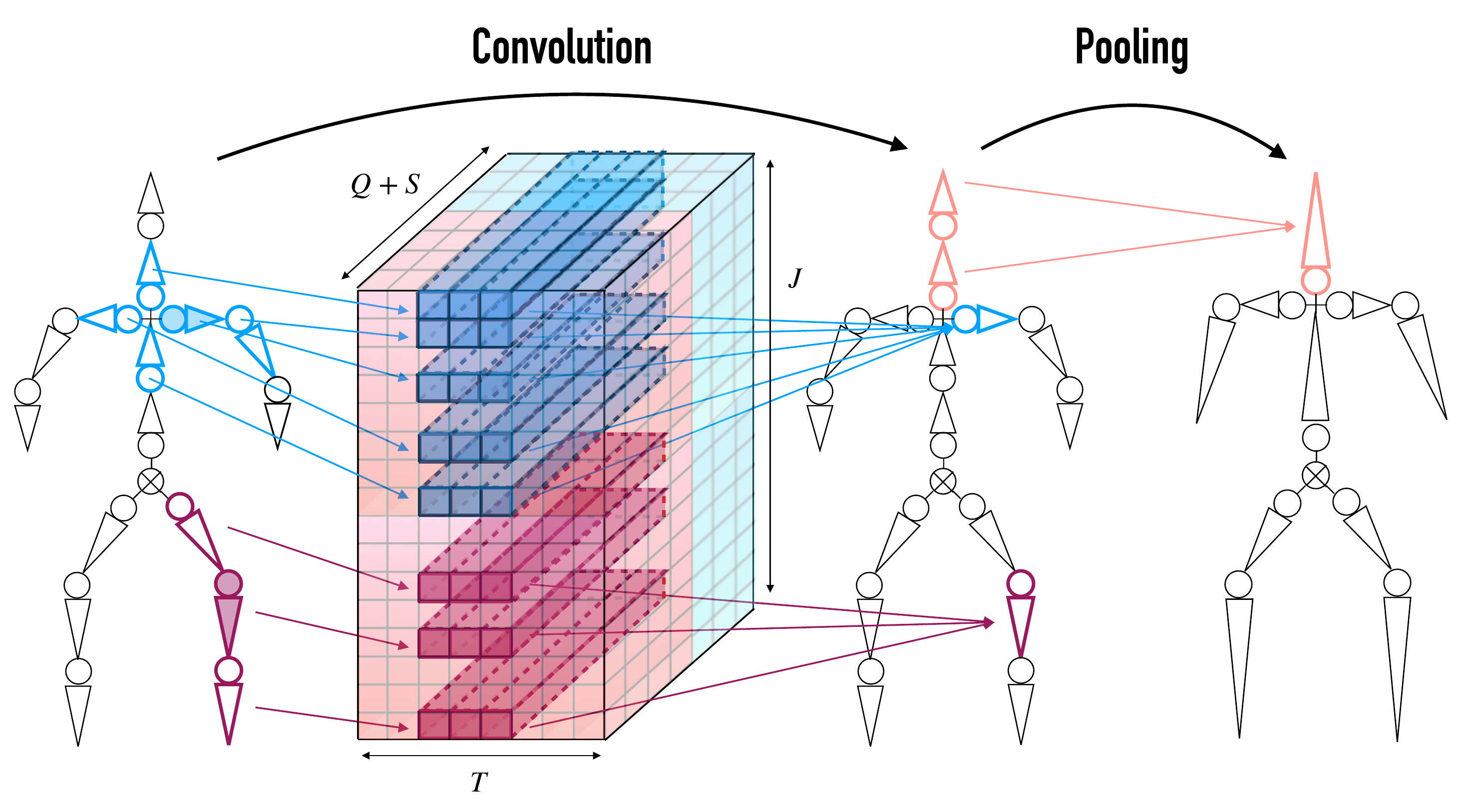} 
	\caption{Skeletal Convolution and Pooling. The skeleto-temporal convolution kernels (e.g., blue and purple) have local support. Support is contiguous along the time axis; for armatures, the support of each kernel is dictated by the connectivity map and the distance (d=1) to the armature at which the kernel is ``centered'' (shown filled in the left skeleton). Weights along the time axis are temporally-invariant, but they are not shared across different armatures. The right skeleton shows the result of topology-preserving skeletal pooling that merges features of pairs of consecutive armatures into single ones.
	\label{fig:skeleton_conv_pool}
	}
\end{figure}

The dynamic branch performs \emph{skeleto-temporal} convolutions, using kernels with local support both along the armature and the temporal axes, as illustrated in Figure~\ref{fig:skeleton_conv_pool}.
Note that while kernels across the time axis are temporally-invariant (namely, the kernel weights are shared along time), they are not shared across different armatures. This is due to the fact that different body parts move in different patterns; thus, the extracted features in each part might be unique. For instance, it is plausible to expect that the features extracted by kernels that are centered at the spine joint might be different than those  centered at the knee. 

Since motion is fully described by a combination of  static and dynamic features, it is important for the convolution kernels to consider both components during computation. Thus, we tile the static representation $\bbs$ along the temporal axis and concatenate the result to $\bbq$ along the channels axis, to yield $\bbm\in\RR^{T\times J\times (Q+S)}$.

In practice, a skeleto-temporal convolution in the dynamic branch  with local support $d$ is applied at every armature via
\begin{equation}
\hat{\bbq}_i = \frac{1}{\vert \mmn_i^d \vert}\sum_{j\in \mmn_i^d}\bbm_j\ast \bbw_j^i + \bb_j^i\,,
\label{eq:dynamic_conv}
\end{equation}
where $\bbm_j\in\RR^{T\times (Q+S)}$ represents the features of the $j$-th armature, $\bbw_j^i\in\RR^{k\times (Q+S) \times K}$ and $\bb_j^i\in\RR^{K}$ are $K$ learned filters with temporal support $k$, and $\ast$ denotes a temporal, one dimensional convolution. Note that the number of armatures before and after convolution is preserved. Figure~\ref{fig:skeleton_conv_pool} illustrates two skeletal convolution kernels (red and blue), where each is applied to a different group of neighbor armatures.

In contrast to the dynamic branch, the static branch takes only the static feature matrix $\bbs$ as input, while ignoring the dynamic part.
This is a choice that we make to ensure that the static components of the resulting deep feature spaces depend only on the structure of the skeleton and not on any particular motion.
Thus, the static convolution operator may be viewed as a special case of the dynamic one in \eqref{eq:dynamic_conv}, with $Q=0$ (i.e., $S$ input channels), $T=1$ and $k=1$. In practice, this is a matrix multiplication operator. 

Both the static and the dynamic branches share the connectivity map ${\bf \mmn}^d$, which enables us to recursively apply skeletal convolution to motion sequences while maintaining dimensional and semantic consistency between the branches.

Note that the root, which is treated as a special armature (with two dynamic parts: global positions and global rotations, is convolved by a kernel whose support contains the closest armatures (up to distance $d$), as well as the end effectors. 
The support is chosen in this manner due to the existing low-level correlation between the global root motion to the local motion of end effectors, as can be clearly observed during running and walking, for example. This connection enables the global information to be injected into the dynamic features of deeper layers.

\subsection{Topology Preserving Skeletal Pooling}
In order to enable our skeleton-aware network to learn higher level, deep skeleto-temporal features, we next define pooling over armatures, which is inspired by the MeshCNN framework \cite{hanocka2019meshcnn}, which merges mesh edges, while pooling their deep features.

In general, pooling encourages the network to learn an efficient basis (kernels) that enables it to extract features of lower dimension, which are optimized to satisfy a specified loss function. For pooling on regular data such as temporal signals or images, adjacency is inherently implied by the signal structure, and the kernel size determines the pooling region. In that case, the features in the corresponding region are merged (usually by averaging or taking the maximum) into a smaller uniform grid, where adjacency is also well-defined.

There are various ways to define pooling on armatures. Our pooling is \emph{topology-preserving}, meaning that the pooled skeleton (graph), which contains fewer armatures (edges), is homeomorphic, to the input one.
Specifically, our pooling operator removes nodes of degree 2, by merging the features of their adjacent edges. This definition is motivated in the next section.
 
Pooling is applied to skeletal branches with a consecutive sequence of edges that connect nodes of degree 2, where the pooling regions are disjoint sets of edges $\{P_1, P_2, \ldots, P_{\tilde{J}}\}$, whose size is not larger than $p$. A sequence of $N$ edges will be split into $\lfloor\frac{N}{p}\rfloor$ sets of $p$ edges and another set of size $N - p\lfloor\frac{N}{p}\rfloor$, in the case that $N$ is not divisible by $p$. We select the remainder set to be the one that is closest to the root. In practice, since  sequential branches in human skeletons are short, we consistently use $p=2$. 

The skeletal pooling is applied to both the static and dynamic feature activations, and formally given by
\begin{equation}
  \hat{\bbs}_i = \text{pool}\{\bbs_{j} \  \vert \ j\in P_i \} \quad\mbox{and} \quad\hat{\bbq}_i= \text{pool}\{\bbq_{j} \ \vert \ j\in P_i \},
\end{equation}
where $\text{pool}$ can be either max or average. The skeletal pooling operator over the armature axis is illustrated in Figure~\ref{fig:skeleton_conv_pool}. It can be seen, for example, that the sequential branch from the neck to the head (marked in red), which contains two armatures is pooled into a single armature in the resulting skeleton. Our pooling can be intuitively interpreted as an operation that enables the network to learn a deep skeleton, with fewer armatures, which approximates the motion of the original skeleton. 
Note that in the dynamic branch a standard downsampling is also additionally applied to the temporal axis.

\paragraph{Unpooling}
The unpooling operator is the counterpart of pooling,  increasing the resolution of the feature activations without increasing the information. 
In our case, unpooling is performed based on the recorded structure of the prior pooled skeleton. We expand the number of edges (armatures) by copying the feature activations of each edge that is originated by a merging of two edges in the corresponding pooling step.
Since unpooling does not have learnable parameters, it is usually combined with convolutions to recover the original resolution lost in the pooling operation.
Note that in the dynamic branch standard upsampling is additionally applied to the temporal axis.

\begin{figure}
	\centering
	\includegraphics[width=\linewidth]{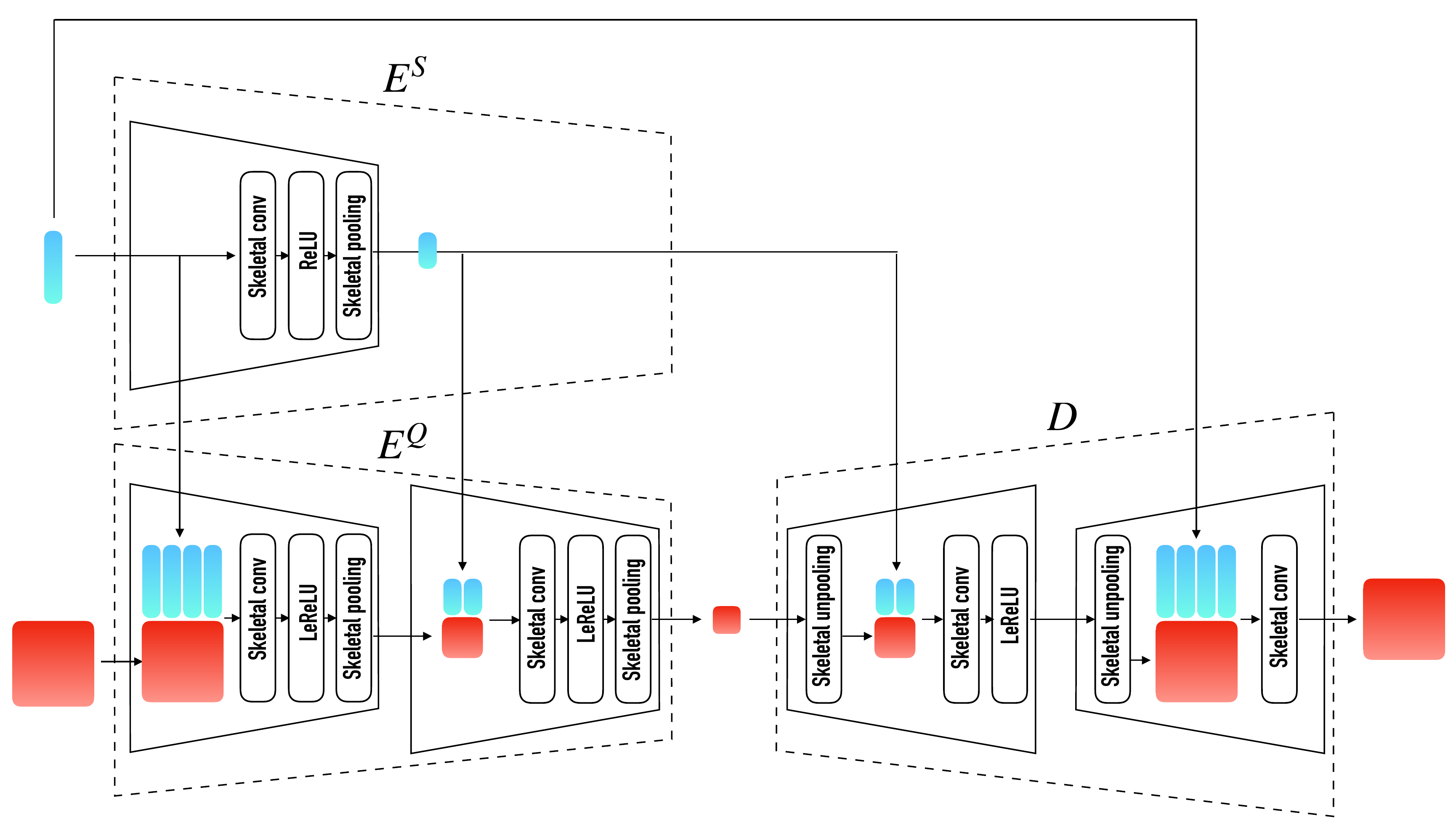} 
	\caption{An autoencoder with two skeletal blocks in the encoder and in the decoder. The static features are encoded and concatenated into both downsampling and upsampling blocks.
	\label{fig:autoencoder}
	}
\end{figure}

\subsection{Evaluation}
\label{sec:denoising}
Our framework can be useful for various learning-based motion processing tasks. We next evaluate the building blocks of our framework against those proposed by Holden~\etal~\shortcite{holden2016deep}, who introduced a deep learning framework for motion editing. Their building blocks consist of standard 1D temporal convolutions, with a full support over the channel (joint) axis and pooling operators which are performed only on the temporal axis. 

In order to evaluate the effectiveness of the two frameworks, we implemented two autoencoders, which share the same number of components and type of layers. In the first, we used the standard convolution and pooling proposed Holden~\etal~\shortcite{holden2016deep}, while in the second we used our skeleton-aware operators. Figure~ \ref{fig:autoencoder} depicts a diagram of our autoencoder, that contains a set of static and dynamic encoders ($E^Q$ and $E^C$, respectively), and a decoder $D$.
Details about the number of input/output channels in each layer are given in the Appendix~\ref{appendix}. 

Both autoencoders are trained with a single reconstruction loss ($\ell_2$ norm), using our dataset described in Section~\ref{sec:experiments}. 

For a fair comparison, both autoencoders were trained on joint rotations, represented by unit quaternions (although in the original paper, Holden~\etal~\shortcite{holden2016deep} use joint positions to represent motion). However, in order to avoid error accumulation along the kinematic chains, the reconstruction loss is applied to the corresponding joint positions, obtained from the rotations by forward kinematics (FK).

During training, each of the autoencoders learns a latent space that represents a motion manifold: a continuous space of natural motions; thus, motion denoising can be performed by simply projecting a noisy motion onto the manifold, using the trained encoder, and then decoding it back to space-time.

We evaluate the performance of the autoencoders by measuring the reconstruction ($\ell_2$ loss) on a test set of unseen motions, where the inputs are injected with two types of noise: (i)  White Gaussian noise ($\mu=0$, $\sigma=0.01$) (2) random zeros: we randomly select pairs of joints and frames and overwrite the existing values with zeros (simulating MoCap glitches).
 
\begin{figure}
	\centering
	\begin{tabularx}{\columnwidth}{XXX}
	\hspace{0.7cm} Input &  \hspace{0.2cm} Holden~\shortcite{holden2016deep} & \hspace{0.8cm} Ours
\end{tabularx}
	\includegraphics[width=\linewidth]{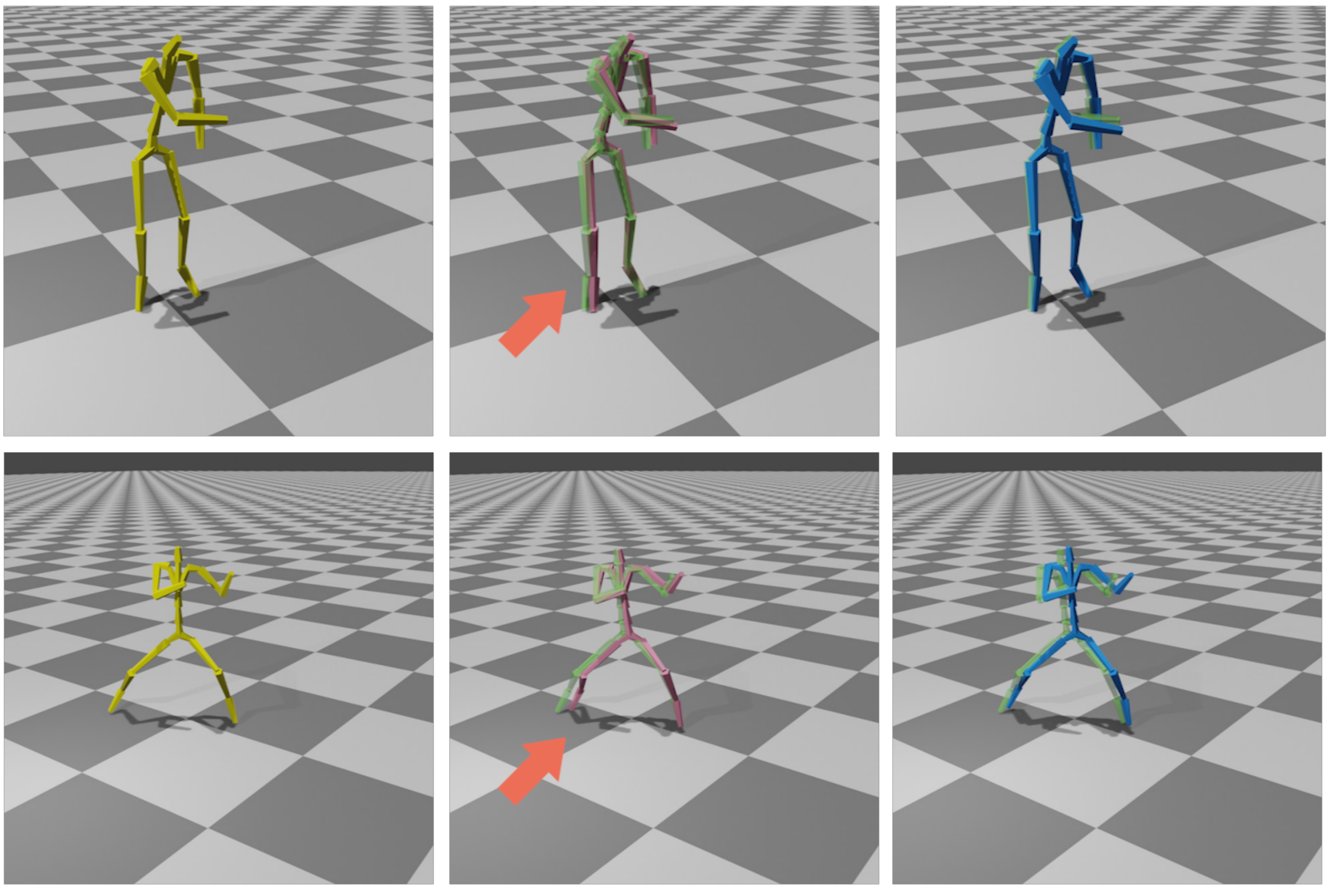} 
	\caption{Motion denoising performance comparison. We compare our skeleton-aware operators to the operators used by Holden~\etal~\shortcite{holden2016deep}. Input (left), Holden~\etal~\shortcite{holden2016deep} (middle), ours (right). Ground truth is overlaid (green skeleton).
	\label{fig:denoising}
	}
\end{figure}

Figure~\ref{fig:denoising} shows frames extracted from the video sequences that can be found in the supplementary material, and Table~\ref{tab:denoising} reports a quantitative comparison between the methods. It can be seen that our skeleton-aware operators achieve better performance for both noise types. As can be observed in the video, our results demonstrate smaller local errors in joint positions and also better global positions, as well as stability. The conventional operators ignore the skeleton structure, while ours pool and compress the information in a more structured manner.

\begin{table}
\caption{Quantitative comparison between our method and that of Holden~\etal~\shortcite{holden2016deep} on a denoising task, where two different types of noise are applied to the inputs. We report the average error over all joints and all motions, normalized by the skeleton's height (multiplied by $10^3$, for clarity).}
\begin{tabular}{l c c}
\toprule
& Holden \etal~\shortcite{holden2016deep} & Ours \\
\midrule
White noise & 1.08 & 0.74\\
Random zeros & 1.08 & 0.81\\
\bottomrule
\end{tabular}
\label{tab:denoising}
\end{table}

\section{Cross-Structural Motion Retargeting}
\label{sec:retargeting}

Motion retargeting is not a precisely defined task.
When a source motion is manually retargeted by animators to a target character, they usually attempt to achieve two main goals: first, the resulting motion should appear natural and visually plausible for the target character, while closely resembling the original source motion.
Second, that joint positions satisfy perceptually sensitive constraints, such as foot and hand contact, typically achieved by applying IK optimization. 
Below, we explain how our framework enables unsupervised retargeting that follows the aforementioned rules. 

\subsection{Problem Setting}
We formulate motion retargeting as an unpaired cross-domain translation task.
Specifically, let $\mm_A \mbox{ and } \mm_B$ denote two motion domains, where the motions in each domain are performed by skeletons with the same skeletal structure ($\mms_A\mbox{ and }\mms_B$, respectively), but may have different bone lengths and proportions. This formulation fits existing public MoCap datasets, where each dataset contains different characters that share the skeletal structure and performing various motions. It is further assumed that a homeomorphism exists between the skeletal structures of $\mms_A$ and $\mms_B$. Note that the domains are unpaired, which means that there are no explicit pairs of motions (performed by different skeletons) across the two domains.

Let each motion $i \in \mm_A$ be represented by the pair $(\bbs_A,\bbq_A^i)$, where $\bbs_A \in \mms_A$ is the set of skeleton offsets and $\bbq_A^i$ are the joint rotations, as described in Section~\ref{subsec:motionrep}.
Given the offsets of a target skeleton $\bbs_B\in\mms_B$, our goal is to map $(\bbs_A,\bbq_A^{i})$ into a retargeted set of rotations $\tilde{\bbq}_B^{i}$ that describe the motion as it should be performed by $\bbs_B$. Formally, we seek a data-driven mapping $G^{A\rightarrow B}$ 
\begin{equation}
G^{A\rightarrow B}\left((\bbs_A,\bbq_A^i)\in\mm_A,\  \bbs_B \in \mms_B\right) \ \ \rightarrow \ \ (\bbs_B,\tilde{\bbq}_B^i)
\end{equation}
In our settings, the translation mapping $G^{A\rightarrow B}$ is learned concurrently with the mapping $G^{B\rightarrow A}$.

\begin{figure}
	\centering
	\includegraphics[width=\linewidth]{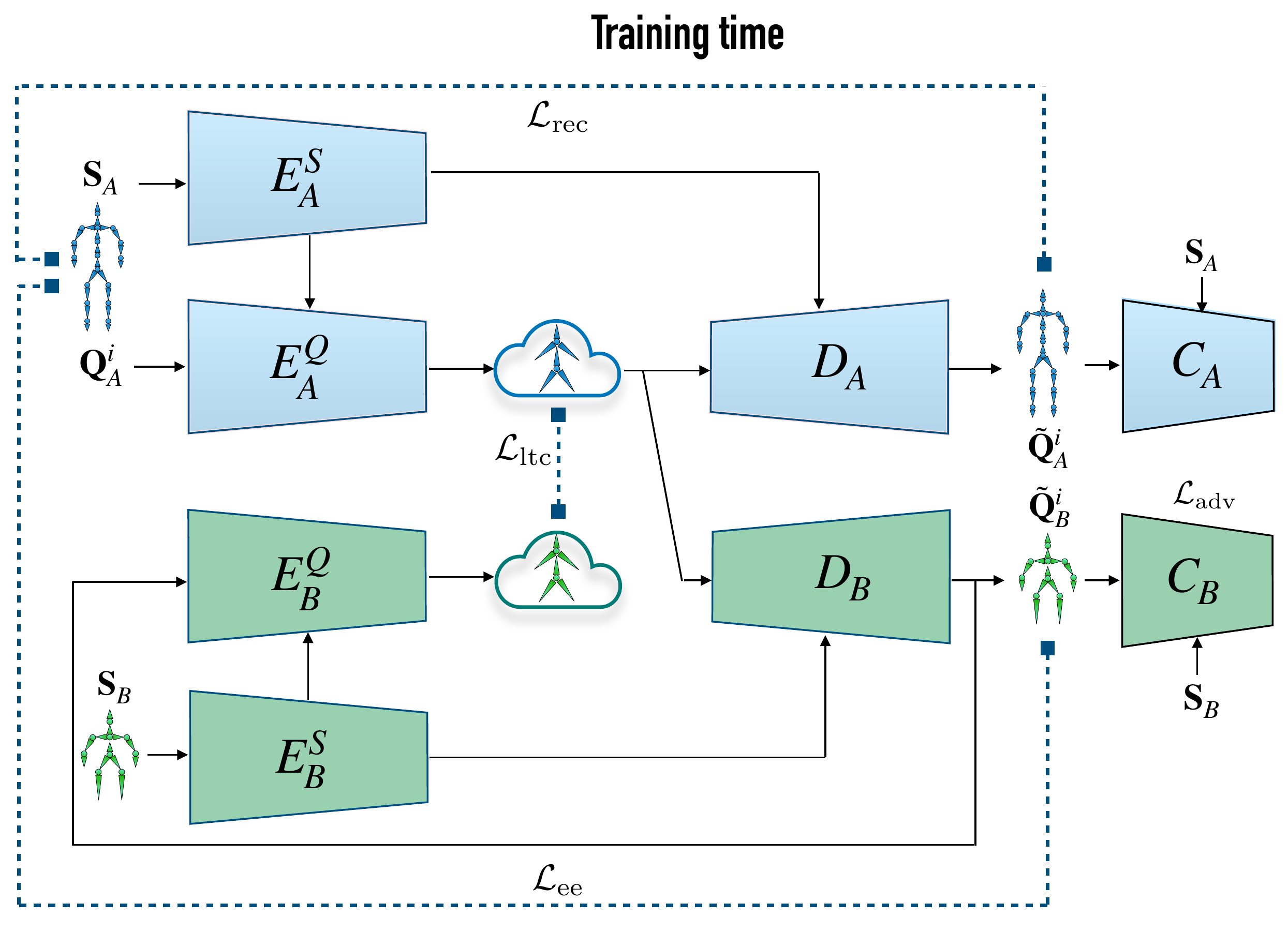} \\ (a) \\
	\includegraphics[width=\linewidth]{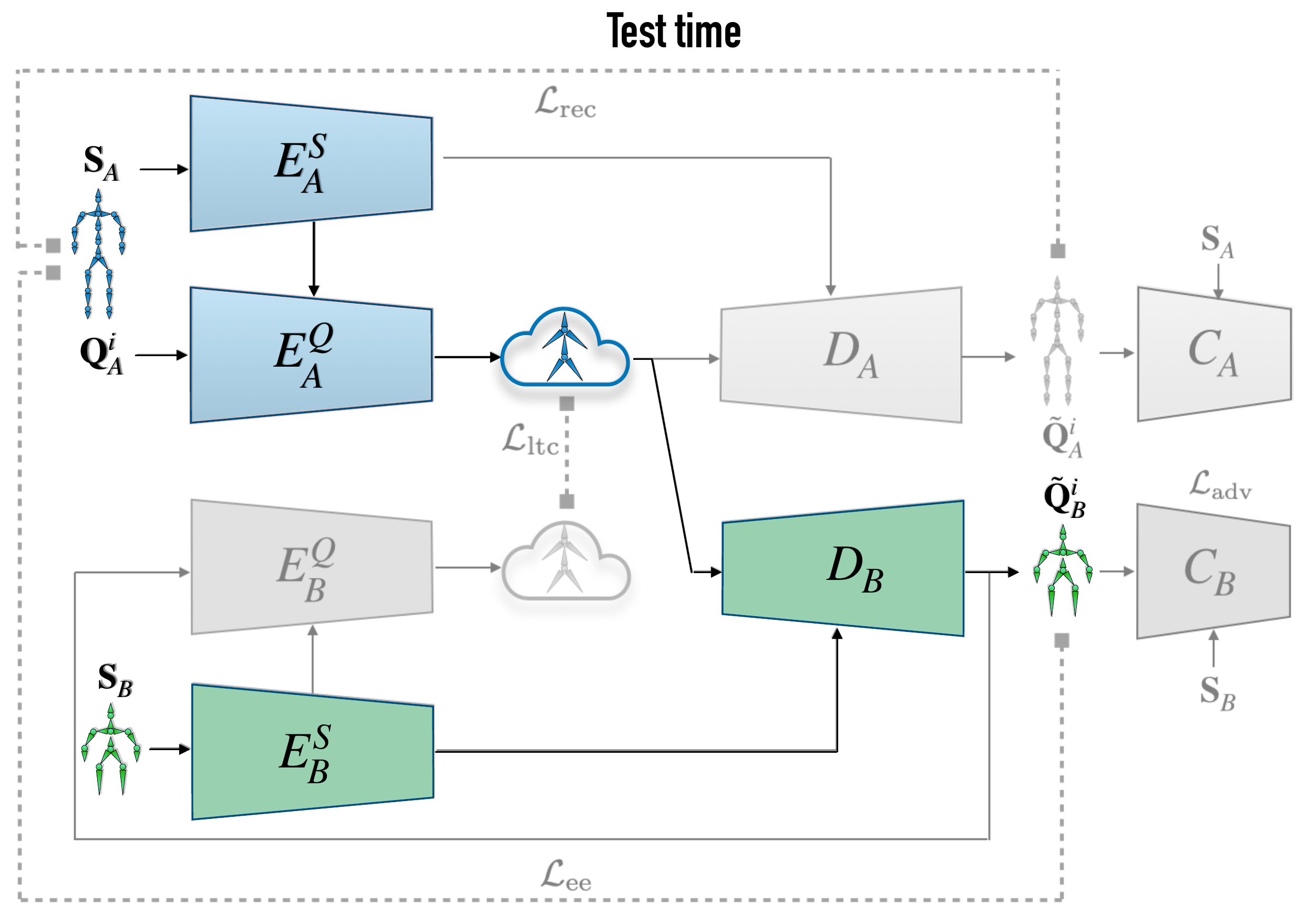} \\ (b) \\
	\caption{Our cross-structural retargeting architecture. (a) Training time: Dynamic and static features from domain $A$, $(\bbs_A,\bbq_A^i)$, are encoded by $E^Q_A$ and $E^S_A$, respectively. Reconstruction by decoder $D_A$ is enforced via $\Loss_{\text{rec}}$. Cross-structural motion translation is achieved by feeding the output of $E^Q_A$ and $E_B^S$ into $D_B$ and applying the  end effectors loss, $\Loss_{\text{ee}}$, along with latent consistency loss $\Loss_{\text{lts}}$ between the original latent representation produced by $E^Q_A$ and that of the translated motion by $E^Q_B$. (b) Test time: Retargeting is performed by using $D_B$ to combine motion encoded by $E_A^Q$ with skeletal features encoded by $E_B^S$. Note that the diagrams describe only $A\rightarrow B$ retargeting, while in practice the training is symmetric.
	}
	\label{fig:retarget_full}
\end{figure}

\subsection{Network Architecture}

Our architecture consists of encoders, decoders and discriminators, with an individual combination of an encoder $E_m=[E_m^Q, E_m^S]$, a decoder $D_m$, and a discriminator $C_m$  is trained for each domain
$\mm_{m}, \ m \in \{A,B\}$. Here, $E_m^Q$ is the dynamic encoder and $E_m^S$ is the static one, as shown in Figure~\ref{fig:autoencoder}. Also, see Figure~\ref{fig:retarget_full}(a), which shows a higher level view of the information flow in our network. 

Having trained the aforementioned components for each motion domain, the desired mapping $G^{A\rightarrow B}$ is obtained, at test time, by using the decoder $D_B$ to combine the dynamic motion representation produced by $E_A^Q$ with the static representation produced by $E_S^B$, as depicted in Figure~\ref{fig:retarget_full}(b).
This is possible, since our encoders produce a deep encoding of motion, which is independent of the original skeletal properties, and the shared latent space is associated with a common primal skeletal structure.
In other words, the encoders disentangle the low-level, correlated dynamic and static parameters, and the retargeting can be simply performed using their deep, disentangled representation.

Below, we describe the different losses used to train our network, also depicted in Figure~\ref{fig:retarget_full}(a). 
For simplicity, we denote the encoded dynamic features by $\bar{\bbq}_A^i=E_A^Q(\bbq_A^i,\bbs_A)$, $\bar{\bbq}_B^i=E_B^Q(\bbq_B^i,\bbs_B)$, and denote the set of static deep features, coupled with the input skeleton, by $\bar{\bbs}_A=\{E_A^S(\bbs_A), \bbs_A\}$, $\bar{\bbs}_B=\{E_B^S(\bbs_B), \bbs_B\}$. The latter notation simplifies the equations since the dynamic encoders and decoders receive the input skeleton as well as a set of deep skeletal features (which are concatenated along their deep layers), as can be seen in Figure~\ref{fig:autoencoder}.
Note that our loss terms are described only for one direction ($A\rightarrow B$). The symmetric term is obtained by swapping the roles of $A$ and $B$.

\paragraph{Reconstruction Loss}
To train an auto encoder $\left([E_A^Q,E_A^S], D_A\right)$ for motions in the same domain, we employ a standard reconstruction loss over the joint rotations and joint positions
\begin{eqnarray}
\Loss_{\text{rec}} & = &  \bbe_{(\bbs_A,\bbq_A^i)\sim \mm_A}\left[ \,\left\|(D_A(\bar{\bbq}_A^i, \bar{\bbs}_A), \bbs_A)-\bbq^{i}_A \right\|^2 \right]\\ & + & \bbe_{(\bbs_A,\bbq_A^i)\sim \mm_A}\left[ \,\left\|\FK(D_A(\bar{\bbq}_A^i, \bar{\bbs}_A), \bbs_A)-\bbp^{i}_A \right\|^2 \right] ,
\end{eqnarray}
where $\FK$ is a forward kinematic operator that returns the joint positions (given rotations and a skeleton) and $\bbp^{i}_A=\FK(\bbq^i_A,\bbs_A)$ are the joint positions of the input character. Note that the positions are normalized by the height (feet to head) of the character. The presence of the loss over the joint positions prevents accumulaiton of error along the kinematic chain \cite{pavllo2019modelingHM}.

\paragraph{Latent Consistency Loss}
As mentioned earlier, our skeletal pooling operator enables embedding motions of homeomorphic skeletons into a common deep primal skeleton latent space, by pooling the features of consecutive armatures, as illustrated in Figure~\ref{fig:pool_to_primal}. 

Embedding samples from different domains in a shared latent space has proved to be efficient for multimodal image translation tasks~\cite{huang2018multimodal, gonzalez2018image}. Constraints may be applied directly on this intermediate representation, facilitating disentanglement.
Inspired by this, we apply a latent consistency loss to the shared representation to ensure that the retargeted motion $\tilde{\bbq}_{B}^{i}$ retains the same dynamic features as the original clip:
\begin{equation}
\Loss_{\text{ltc}}  = \bbe_{(\bbs_A,\bbq_A^i)\sim \mm_A}\left[ \,\left\| E_B^Q(\tilde{\bbq}_{B}^{i},\bar{\bbs}_B)-E_A^Q(\bbq^{i}_A,\bar{\bbs}_A) \right\|_1 \right],
\end{equation}
where $\Vert \cdot \Vert_1$ is the $L_1$ norm.

\paragraph{Adversarial Loss}
Since our data is unpaired, the retargeted motion has no ground truth to be compared with. Thus, we use an adversarial loss, where a discriminator $C_B$ assesses whether or not the decoded temporal set of rotations $\tilde{\bbq}_B^{i}$ appears to be a plausible motion for skeleton $\bbs_B$:
\begin{equation}
\Loss_{\text{adv}}  = \bbe_{i\sim \mm_A}\left[ \Vert C_B(\tilde{\bbq}_B^{i}, \bar{\bbs}_B) \Vert^2 \right] + \bbe_{j\sim \mm_B}\left[ \Vert 1-C_B(\bbq_B^{j}, \bar{\bbs}_B)\Vert^2 \right].
\end{equation}
As in other generative adversarial networks, the discriminator $C_B$ is trained using the motions in $\mm_B$ as the real examples, and the output of $G^{A\rightarrow B}$ as the fake ones.

\paragraph{End-Effectors Loss}
While homeomorphic skeletons may differ in the number of joints, they share the same set of end-effectors. We exploit this property to require that the end-effectors of the original and the retargeted skeleton have the same normalized velocity. The normalization is required since velocities may be at different scales for different characters.
This requirement is particularly helpful to avoid common retargeting artifacts, such as foot sliding; frames with zero foot velocity in the input motion should result in zero foot velocity in the retargeted motion. This is formulated by
\begin{equation}
\Loss_{\text{ee}}  =  \bbe_{i\sim \mm_A}\sum_{e\in \mathcal{E}} \left\Vert \frac{ V_{A_e}^{i}}{h_{A_e}}- \frac{ V_{B_e}^{i}}{h_{B_e}} \right\Vert^2 ,
\end{equation}
where $V_{A_e}^i$ and $V_{B_e}^i$ are the magnitudes of the velocity of the $e$-th end-effector of skeletons $\bbs_A$ and $\bbs_B$, respectively, while performing motion $i$,  $\mathcal{E}$ is the set of end-effectors, and $h_{A_e}, h_{B_e}$ are the lengths of the kinematic chains from the root to the end-effector $e$, in each of the skeletons $\bbs_A$ and $\bbs_B$.

Although our end-effectors loss significantly mitigates foot sliding artifacts, we further clean foot contact using standard Inverse Kinematics (IK) optimization.
The cleanup is fully automatic: we extract binary foot contact labels from the motion input sequence, and apply IK to enforce foot contact by fixing the position of the foot to the average position along contact time slots. The effects of our end-effectors loss and the IK-based cleanup, are demonstrated in supplementary video.

The full loss used for training combines the above loss terms: 
\begin{equation}
\Loss = \Loss_{\text{rec}} + \lambda_{\text{ltc}}\Loss_{\text{ltc}} + \lambda_{\text{adv}}\Loss_{\text{adv}} + \lambda_{\text{ee}}\Loss_{\text{ee}}
\label{eq:tot_loss}
\end{equation}
where  $\lambda_{\text{ltc}}=1$, $\lambda_{\text{adv}}=0.25$, $\lambda_{\text{ee}}=2$. 

\section{Experiments and Evaluations}
\label{sec:experiments}
In this section we evaluate our results, compare them to other retargeting methods, and demonstrate the efficiency of various components in our framework. In order to qualitatively evaluate our results, please refer to the supplementary video.

\subsection{Implementation Details}
Our motion processing framework is implemented in PyTorch, and the experiments are performed on a PC equipped by an NVIDIA GeForce GTX Titan Xp GPU (12 GB) and Intel Core i7-695X/3.0GHz
CPU (16 GB RAM). We optimize the parameters of our network, with the loss term in \eqref{eq:tot_loss}, using the Adam optimizer \cite{kingma2014adam}.
Training our skeleton-aware network takes about 22 hours (around 5000 epochs), however, the performance can be improved with a parallel implementation of our operators. 

In order to evaluate our method, we construct a dataset with 2400 motion sequences, performed by 29 distinct characters, from the Mixamo 3D characters collection \cite{mixamo}.
For each motion, we randomly choose a single character to perform it, to ensure that our dataset contains no motion pairs. Furthermore, to enable training the network in batches, motions are trimmed into fixed temporal windows with $T=64$ frames each. However, note that our network is fully-convolutional, so there is no limitation on the length of the temporal sequence at test time.  

The characters can be divided into two groups, $A$ and $B$, each containing skeletons with similar structure but different body proportions. 
This configuration fits our problem setting and also enables us to quantitatively evaluate our method against a ground truth. 
Note that due to imbalance in the Mixamo character collection, group $A$ contains 24 characters, while group $B$ contains only 5 (the characters in group $B$ are the only characters with this skeletal structure in the original dataset). 
Our framework assumes that all of the characters contain 5 main limbs (2 hands, 2 feet, and head), thus, we clip the fingers of the Mixamo characters.

\subsection{Intra-Structural Retargeting}
Our framework can be used also for retargeting of skeletons with the same structure, but different proportions, where a single translator (encoder-decoder) is used to perform the task. In this experiment we compare our method to other methods that perform intra-structural retargeting. We use group $A$ to train the network on 20 characters and test it on 4 (unseen) ones. 

The first method we compare to is Neural Kinematic Networks (NKN) of Villegas~\etal~\shortcite{villegas2018neural},  a recurrent neural network architecture with a Forward Kinematics layer to cope with intra-structural motion retargeting. Another state-of-the-art method we compare to is PMnet, by Lim~\etal~\shortcite{lim2019PMnetLO}, which uses a two-branch CNN to perform unsupervised motion retargeting for skeletons with similar structures. In addition, since retargeting in our settings corresponds to an unpaired translation, we also compare to a naive adaptation of CycleGAN~\cite{zhu2017unpaired} to the motion domain. In this implementation, each of the domains contains a set of motions that are performed by a distinct character. Two generators and two discriminators (each consisting of a set of 1D temporal convolutions) are trained to translate motions between two domains. An adversarial loss and a cycle consistency loss are used to perform the training. We train such a system for each pair of characters (domains). Each motion is represented as a temporal set of joint rotations (static information is not considered, since there is a single skeleton in each domain). Note that (by definition) the characters are seen, thus, no unseen evaluation is performed on this network. The full implementation details are described in the Appendix~\ref{appendix}. 

Figure~\ref{fig:same_structure_comparison} shows a few frames from the comparison, which is included in the supplementary video. It can be seen that our results (right column) are more stable than those of the other methods and aligned better with the ground truth (a green skeleton, overlaid on top of each output), in terms of global positioning and local errors of joints. 

Another naive baseline for motion retargeting is obtained by simply using the unmodified joint rotations of the source motion with the target skeleton. In practice, in IK based intra-structure retargeting, the source rotations are used to initialize the optimized rotation values, which are then tuned based on contact constraints that depends on the content of the motion. 
In the cross-structural setting, where there is no one-to-one joint correspondence, the mapping is defined manually, which causes unavoidable retargeting errors (due to missing corresponding joints), that should be manually corrected. 

Since in both cases, manual intervention is required (specifying motion-dependent constraints and/or joint correspondence) this baseline is not scalable. Thus, we use a naive copy of rotations as a baseline instead, only for the intra-structural retargeting scenario. 

Table~\ref{tab:quant_comparison} presents a quantitative evaluation of the different methods described above.
Errors are measured by performing retargeting on each pair in the test set (6 pairs totally) over all the $N=106$ test motions and comparing to the ground truth, available from the original Mixamo dataset.
Since (in the cross-structural case) we may deal with a different number of joints (per domain) we calculate the retargeting error of every target skeleton $k$, to all the others in the test set $\mathcal{C}$, as the average distance between joint positions, which is given by
\begin{equation}
E^k = \frac{1}{NJ(|\mathcal{C}|-1)h_k}\sum_{c\in \mathcal{C}, c\neq k}\sum_{i=1}^{N}\sum_{j=1}^{J} \Vert \tilde{\bbp}^j_{ik,c} - \bbp^j_{ik} \Vert,
\end{equation}
where $\tilde{\bbp}^j_{ik,c}$ denotes the $j$-th joint position of the retargeted motion sequence $i$ that was originally performed by skeleton $c$ and transferred to skeleton $k$, $h_k$ is the height of skeleton $k$ and $\bbp^j_{ik}$ is the ground truth. The final error is calculated by averaging the errors $E_k$ for all the skeletons $k\in\mathcal{C}$. The results reported in Table~\ref{tab:quant_comparison} demonstrate that our method outperforms the other approaches in intra-structural retargeting.

\begin{figure*}
	\centering
\begin{tabularx}{\textwidth}{XXXX}
	\hspace{2.8cm}Input &  \hspace{1.3cm}CycleGAN~\shortcite{zhu2017unpaired}  & \hspace{1.0cm}NKN~\shortcite{villegas2018neural} & \hspace{0.65cm}Ours
\end{tabularx}
	\includegraphics[width=0.85\linewidth]{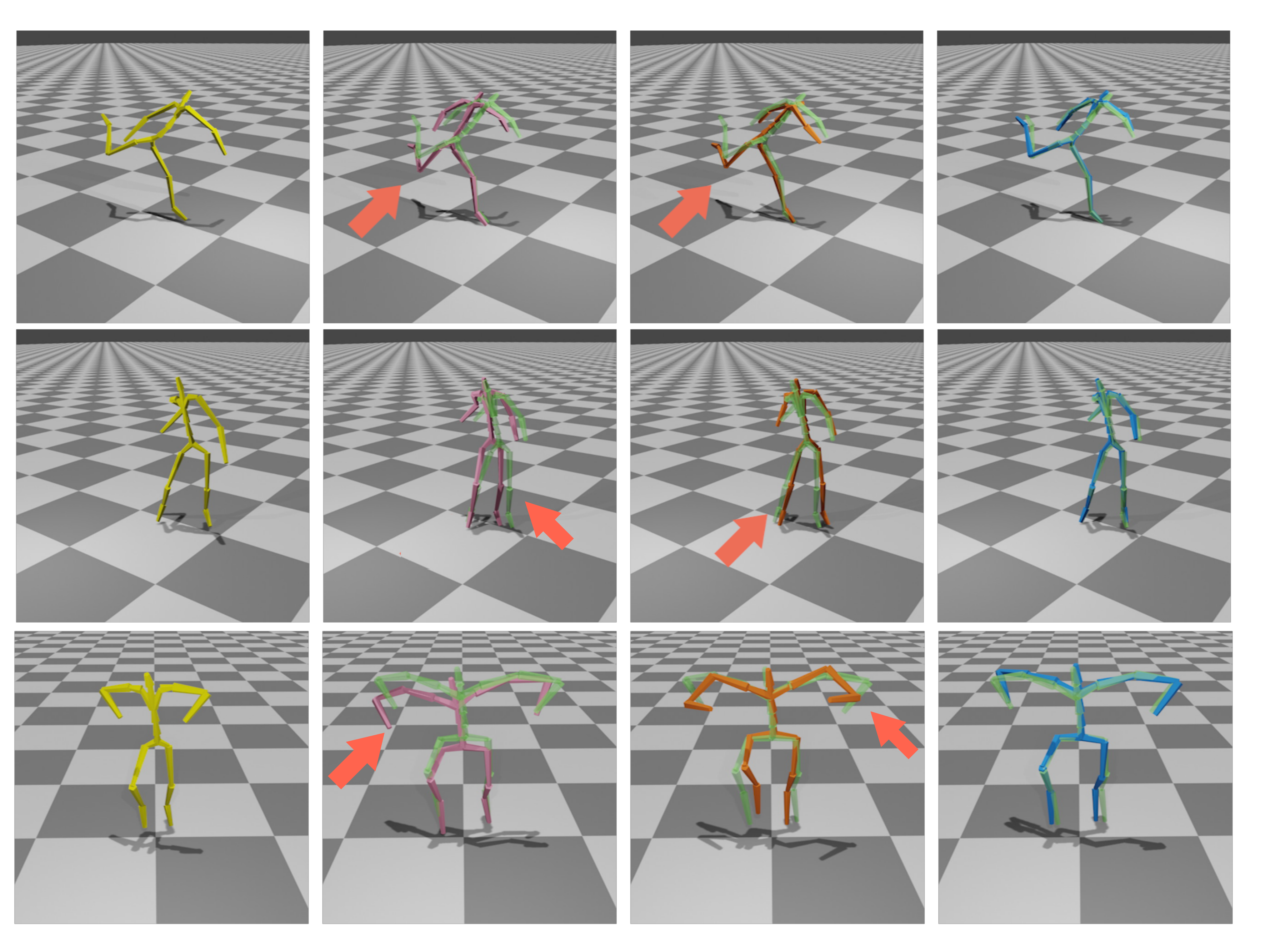} 
	\caption{Intra-structure motion retargeting. Our method is compared to a naive adaptation of CycleGAN~\cite{zhu2017unpaired} to the motion domain and to NKN of Villegas~\etal~\shortcite{villegas2018neural}. The outputs are overlaid with the ground truth (green skeleton). It can be seen that our results are more stable and better aligned with the ground truth. The full comparison can be seen in the supplementary video.}
	\label{fig:same_structure_comparison}
\end{figure*}

\begin{table}
\caption{Quantitative comparison between our method to the method of Villegas~\etal~\shortcite{villegas2018neural}, a naive adaptation of CycleGAN~\shortcite{zhu2017unpaired} to unpaired motion translation, and an ablation study to evaluate the various components and loss terms in our framework.
We report the average error over all joints and all motions, normalized by the skeleton's height (multiplied by $10^3$, for clarity).}
\begin{tabular}{l c c}
\toprule
& \small Intra-Structural &  \small  Cross-Structural \\
\midrule
\small  Villegas~\etal~\shortcite{villegas2018neural} (NKN) & 6.24 & 243\\
\small Lim~\etal~\shortcite{lim2019PMnetLO} (PMnet)& 5.72 & N/A\\
\small  CycleGAN~\shortcite{zhu2017unpaired} adaptation & 7.66 & 8.97\\
\small  Copy rotations  & 8.86 & N/A \\
\small Ours - conventional operators  & 3.95 & 3.56\\
\small Ours - no shared latent space & 3.01 & 3.06\\
\small Ours - no $\Loss_{\text{adv}}$ & \textbf{0.47} & 3.81\\
\small Ours - full approach & 2.76 & \textbf{2.25} \\
\bottomrule
\end{tabular}
\label{tab:quant_comparison}
\end{table}

\subsection{Cross-Structural Retargeting}
We next compare our method in a cross-structural setting, between the two groups $A$ and $B$. In practice, group $B$ originally contains an extra joint in each leg, and another one in the neck (compared to $A$), and we further increase the existing difference by adding more joints (one in the spine, and one in each arm), by splitting each of the affected offsets into two equal parts.
Since group $B$ contains only 5 characters we use 4 of them to train the model and only 1 to test, while in group $A$, we keep the same setup.

Since there are no existing methods that can cope with the task of cross-strtucral retargeting in a fully automatic manner, we naively adapt previous approaches to support a cross-structural setup. Here, we use again the CycleGAN~\cite{zhu2017unpaired} motion translator, but modify the number of input and output channels in each translator to match the number of joints in each domain, as well as the number of input channels of the discriminators accordingly.

In the case of NKN \cite{villegas2018neural}, which uses same-structure skeletons (single domain), we modified the original implementation, to support two different domains.
However, since the number of joints is different between the domains, NKN's reconstruction loss had to be removed. In addition, since the domains have different dimensions, the two networks ($A\rightarrow B$, $A\rightarrow B$) cannot share weights, so they had to be trained separately.

Both the qualitative results in the supplementary video (see extracted frames in Figure~\ref{fig:cross_structure_comparison}), and the quantitative comparison in Table~\ref{tab:quant_comparison}, show that our method outperforms these two alternatives. It can be seen that our cross-structural error is, in fact, lower than the intra-structural one, which might come as a surprise; this is due to the fact that the set in group $B$ is much smaller, with smaller differences in their body proportions, making the single test character closer to the ones seen during training. 

As our comparison shows, the reconstruction loss in NKN \cite{villegas2018neural} plays a key role in the good performance that the original system achieves for intra-structural retargeting, and without this loss term, the quality of their results is significantly degraded.

\begin{figure*}
	\centering
	\begin{tabularx}{\textwidth}{XXXX}
		\hspace{2.7cm}Input &  \hspace{1.2cm}CycleGAN~\shortcite{zhu2017unpaired}  & \hspace{1.0cm}NKN~\shortcite{villegas2018neural} & \hspace{0.75cm}Ours
	\end{tabularx}
	\includegraphics[width=0.85\linewidth]{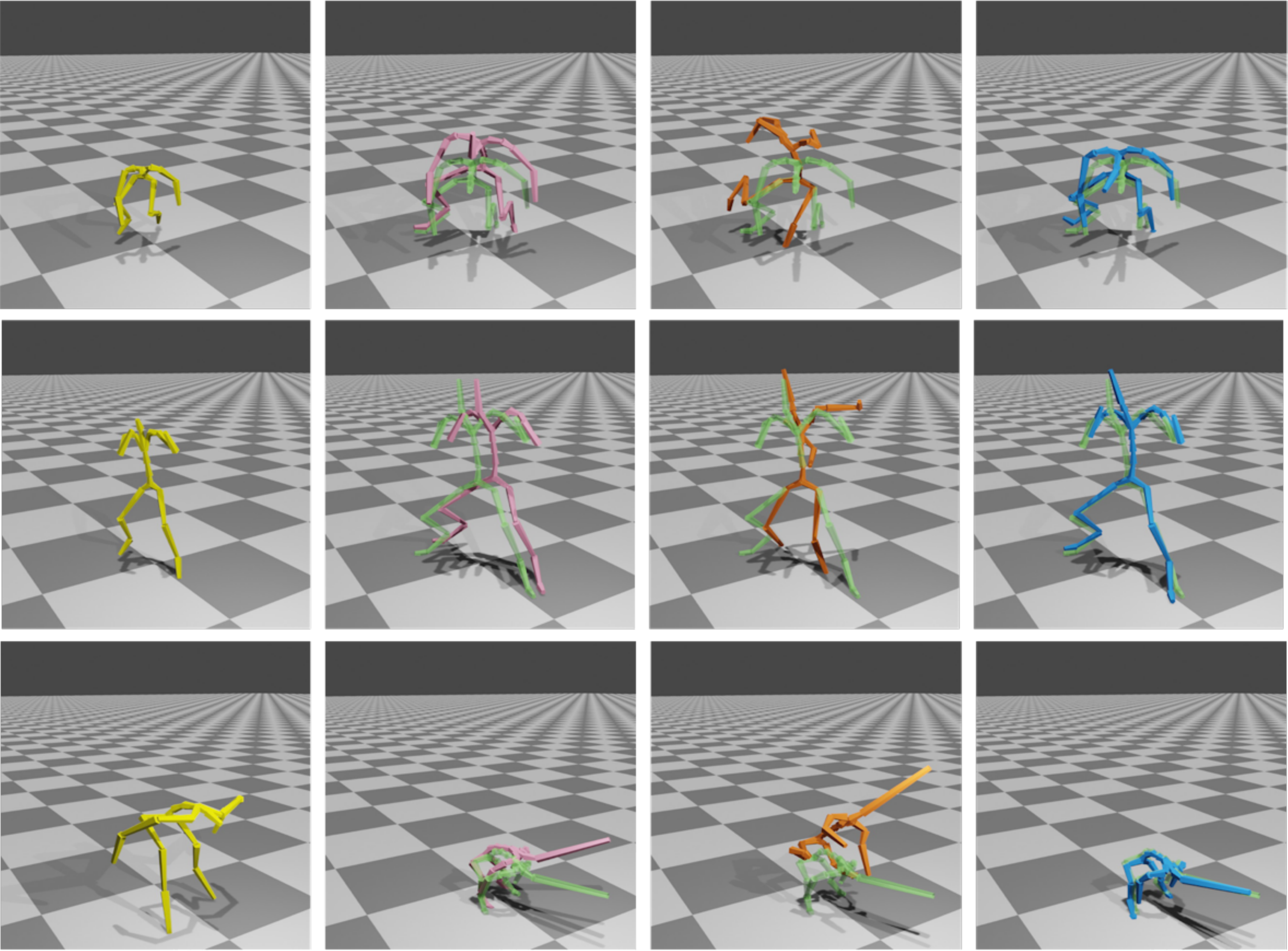} 
	\caption{Cross-structure motion retargeting. Our method is compared to a naive adaptation of CycleGAN~\cite{zhu2017unpaired} to the motion domain and to a cross-structural version of NKN of Villegas~\etal~\shortcite{villegas2018neural}. 
	The outputs are overlaid with the ground truth (green skeleton). The full comparison can be seen in the supplementary video.
	\label{fig:cross_structure_comparison}
	}
\end{figure*}

\begin{figure}
	\centering
	\begin{tabularx}{\columnwidth}{XX}
		\hspace{1.4cm} Input &  \hspace{1.6cm} Ours
	\end{tabularx}

	\includegraphics[width=\linewidth]{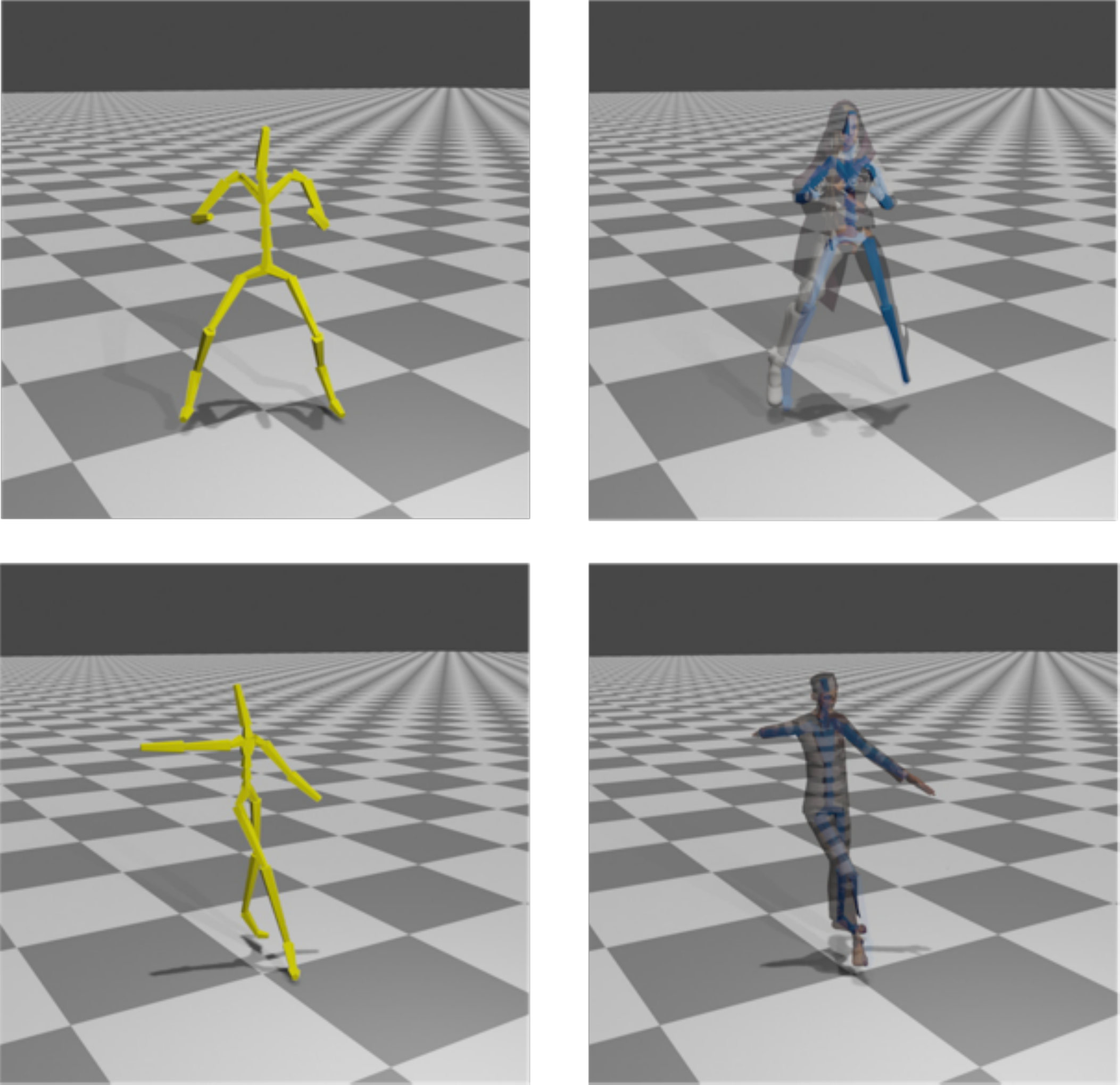} 
	\caption{Retargeting to asymmetric characters. This result shows that our method can deal with asymmetric characters and generates natural results. The full comparison can be seen in the supplementary video.
	\label{fig:asymmetric_characters}
	}
\end{figure}

\begin{figure}
	\centering
	\begin{tabularx}{\columnwidth}{XXX}
		\hspace{0.8cm} Input &  \hspace{0.2cm} Copy rotations & \hspace{0.8cm} Ours
	\end{tabularx}
	\includegraphics[width=\linewidth]{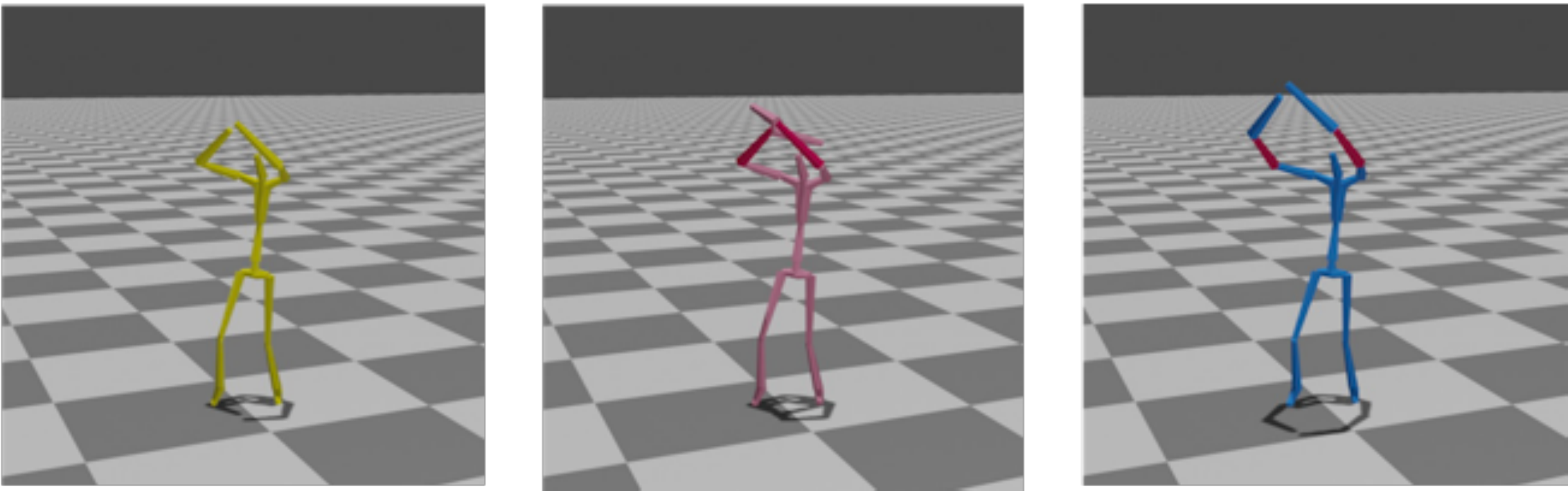} 
	\caption{Retargeting to a character with three bones per arm. Our method is compared to a naive nearest neighbor rotation copying. It can be seen that copying rotations causes implausible results, especially for fine motions like clapping.
	The full comparison can be seen in the supplementary video.
	\label{fig:three_arms}
	}
\end{figure}

\paragraph{Special Characters}
In the supplementary video we demonstrate a few examples of characters with special properties, like asymmetric limbs (missing bone in the leg or arm, see Figure~\ref{fig:asymmetric_characters}), and a character with an extra bone in each arm (see Figure~\ref{fig:three_arms}). Since each of them has a unique skeletal structure, no other skeletons belong to their domain. Thus, in order to perform motion retargeting to other skeletons, we train a special model for each of the examples which translates skeletons from group $A$ to the specific skeleton (and vice versa). It can be seen that our model can learn to retarget skeletons with various structures, as long as they are homeomorphic, including non-trivial motions like clapping hands (see Figure~\ref{fig:three_arms}), which is a very challenging motion to retarget, especially when the lengths of the arms of the source and target skeletons are different. In general, in order to retarget such a motion, the user must manually specify the contact time-positions, and to solve an IK optimization with rotations that are initialized to the rotations of the source character. In the example that appears in the supplementary video, it may be seen that the initial solution achieved by copying the rotations is not satisfactory (due to a missing joint correspondence). 

\subsection{Ablation Study}
In this part we evaluate the impact on performance of our operators, shared latent space, and the various loss terms. The results are reported in Table~\ref{tab:quant_comparison} and demonstrated some parts in the supplementary video.

\paragraph{Skeleton-Aware Operators}
To evaluate the effectiveness of our skeleton-aware operators within the retargeting framework, we perform a comparison where all of them are replaced by conventional ones. The skeletal-convolution is replaced by a standard 1D convolution with a full support over the channels (joints) axis, and the skeletal pooling and unpooling are discarded and upsampling and downsampling is performed only on the temporal axis. In order to still support the structure of a shared latent space, we simply modify the number of output and input channels in each encoder and decoder, respectively, such that both autoencoders share the same latent space size, which equals to the one in the original system (see Appendix~\ref{appendix} for more details).

As reported in Table~\ref{tab:quant_comparison}, although the conventional operators are inferior to the skeleton-aware ones, they still outperform the baselines.
We attribute this to the fact that in our approach the latent space is structured. Thanks to the local armature support of our skeletal convolutions, different kinematic chains are encoded into distinct parts in the latent representation, from which they can be decoded into the target chains by different decoders. For example, an arm with 2 bones and an arm with 3-bones, will be encoded into the same set of channels in the shared representation, and decoded by the relevant encoder.

\paragraph{Shared Latent Space}
In this experiment we retrain our framework without a shared latent space. This is done by replacing the latent consistency loss $\Loss_{\text{ltc}}$ with a full cycle consistency loss on the input. The results in Table~\ref{tab:quant_comparison} show that the performance of this variant is worse.
Without a shared latent space the encoder-decoder can be interpreted as a general translator that translate motion from one domain to another. While this might work well enough for unimodal tasks (where each source point can be mapped into a single target point), multimodal tasks stand more to gain from having a shared latent domain, as already demonstrated in multimodal image translation tasks, e.g., \cite{huang2018multimodal}. 

\paragraph{Adversarial Loss}
In this experiment we discard the adversarial loss $\Loss_{\text{adv}}$ and retrain our network. It can be seen that in the intra-structural setting omitting the adversarial loss actually improves performance, while in the cross-structural retargeting this is not the case. The main reason, in our opinion, is that intra-structural retargeting is a simpler task, where reasonable results may be achieved by copying the rotations and constraining end-effectors. Similarly, the network can easily learn a simple mapping that copies the input rotation values and tunes them using the end-effectors loss, and achieve good results in this manner.

In contrast, in a cross-structural setting, when the joint correspondence between the two domains is not well defined, $\Loss_{\text{adv}}$ is necessary to assist the network in learning how to transfer motions between the two domains.

\paragraph{End-Effectors Loss}
In this experiment we show the effectiveness of our end-effectors loss, by discarding $\Loss_{\text{ee}}$ and retraining our network. 
It is well known that deep networks for character animation, which are not physically-based may output motions with severe foot-sliding artifacts~\cite{holden2016deep}, where the main cause is the natural smoothness of the temporal filters and rotation errors that are accumulated along the kinematic chains. 

Without $\Loss_{\text{ee}}$, when no special attention is given to end-effectors, artifacts can be clearly observed in the outputs (see the supplementary video). The foot positions at contact times are characterized by large errors, which appear like slides. However, when the model is trained with our end-effectors loss, the large errors are mitigated and converted to subtle motions with high-frequency that are concentrated at a fixed point during contact time. These errors, which are less perceptually noticeable, can be easily fixed using a simple, fully-automated, IK based foot-sliding cleanup, as described in Section~\ref{sec:retargeting} and demonstrated in the supplementary video. In addition, Figure~\ref{fig:ablation_ik} shows the magnitude of the velocity of the right foot as a function of time in the motion clip used for the ablation study of $\Loss_{\text{ee}}$ (can be found in the supplementary video 05:18-05:22). It may be seen that without IK (green), the magnitude of the velocity is small but still noisy, characterized by high-frequencies. However, after performing IK (orange), the output is more correlated with the ground truth (blue) and cleaner zero velocity slots can be observed during contact periods.

\begin{figure}
	\centering
	\includegraphics[width=0.9\linewidth]{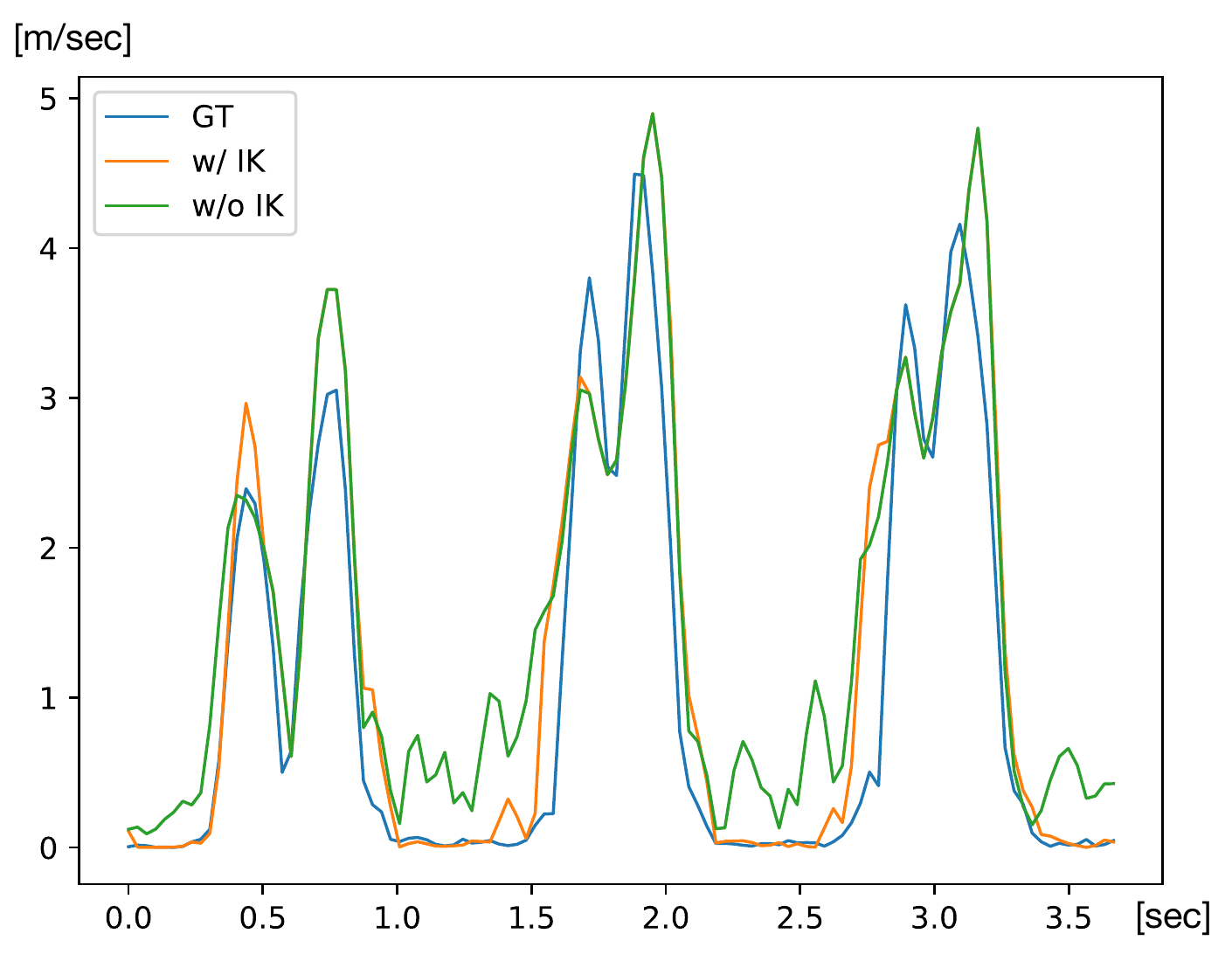} 
	\caption{Foot contact cleanup using IK. The magnitude of the foot velocity in the raw output of the network (green) and the post processed motion signal (orange) are compared to the one in the GT (blue).}
	\label{fig:ablation_ik}
\end{figure}

\section{Discussion and Future work}
\label{sec:discussion}

We have presented a framework where networks are trained to encode sequences of animations of homeomorphic skeletons to a common latent space, and to decode them back, effectively allowing transferring motions between the different skeletons. The success of defining a common encoding space for different skeletons is attributed to three reasons: (i) the primal skeleton is topologically close to the original source skeletons, (ii) it carries enough deep geometric features, which is weakly decoupled from the source skeleton and (iii) we use spatially-variant kernels for convolving the appropriate joint neighborhoods.

\begin{figure}
	\centering
	\begin{tabularx}{\columnwidth}{XX}
		\hspace{1.3cm} Dog T-pose &  \hspace{0.8cm} Gorilla T-pose \\
	\end{tabularx}
	\includegraphics[width=0.4\linewidth]{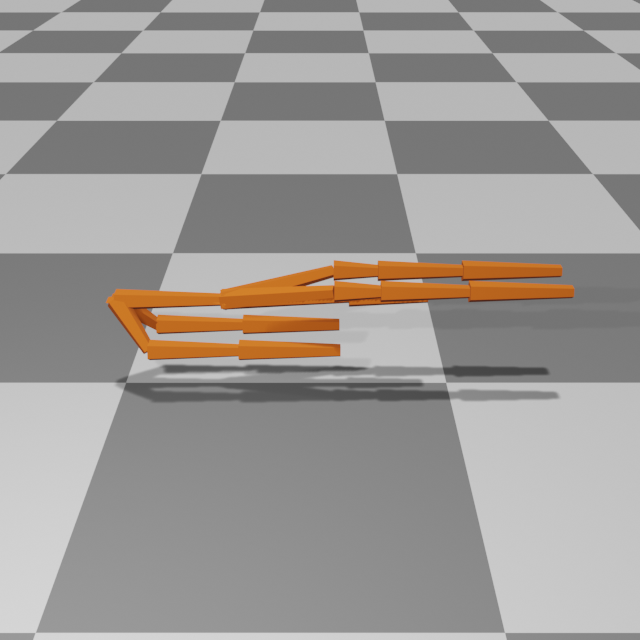} \hspace{3mm}
	\includegraphics[width=0.4\linewidth]{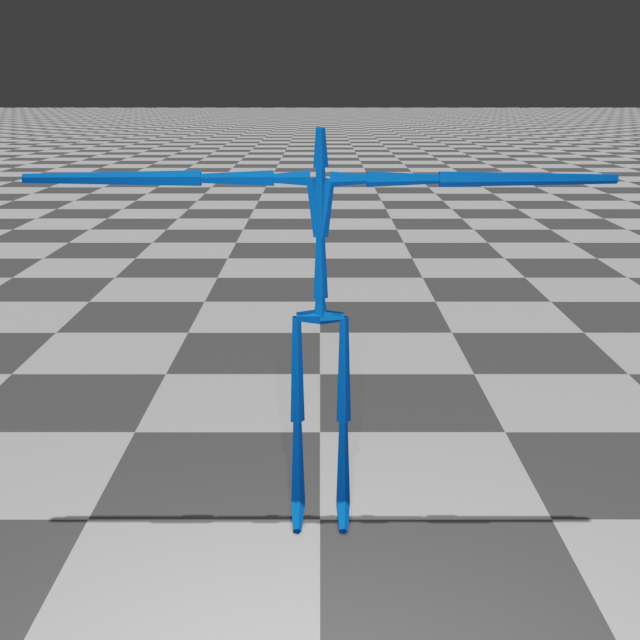}	
	\caption{Drastically different T-poses may hinder retargeting. See the resulting failure example in the supplementary video.
		\label{fig:tposes}
	}
\end{figure}

It is tempting to believe that the primal skeleton could have been reduced down to a single point, and hence potentially allowing to transfer motion among skeletons that are not necessarily homeomorphic. However, our preliminary experiments in that direction suggest that it is too hard to encapsulate such a large amount of information. Nevertheless, extending the scope of compatible skeletons is an interesting topic for future work.


A legitimate question is commonly asked: ``does this really require a deep network solution?'', or ``what does the network learn here''? The answer is that, yes, the deep features that are encoded on the primal skeleton are learned. They are learned by training numerous sequences. The features required to encode an animation decoupled from its joints are overly complex to be modeled by design. Moreover, the deep features of the primal skeleton encode also the bones lengths. This compensates proportion difference between the source and the target skeletons. In other words, motion transfer applied by the network translates both the topology and geometry jointly. Professionals can map motions between pairs of skeletons, however, not only that it is a tedious task, but it often introduces errors that need to manually amended. On the other hand, training networks once, facilitates the transfer between pair of arbitrary sequences from the trained domain.

In the future, we would like to develop means that allows transfer between skeletons that are not homeomorphic. This possibly would require to define a generalized primal skeleton where the topology is disentangled and encoded separately.

A notable limitation is that we cannot retarget motion well between homeomorphic skeletons, if they move very differently from each other and have different T-poses. One such example, attempting to retarget the motion from a dog to a gorilla is included in the supplementary video. Because the joint rotations describing the motion are specified with respect to the T-pose, having very different T-poses, as shown in Figure~\ref{fig:tposes}, implies a completely different interpretation of the joint rotations, making the translation task much more challenging. Moreover, despite our use of the end-effectors loss, complex tasks that involve interactions with objects (like picking up a cup) will not be retargeted properly, if they were not seen in the dataset.

\begin{acks}
We thank the anonymous reviewers for their constructive comments. This work was supported in part by  National Key R\&D Program of China (2018YFB1403900, 2019YFF0302902), and by the Israel Science Foundation (grant no.~2366/16).
\end{acks}

\bibliographystyle{ACM-Reference-Format}
\bibliography{main} 


\begin{thebibliography}{32}


\ifx \showCODEN    \undefined \def \showCODEN     #1{\unskip}     \fi
\ifx \showDOI      \undefined \def \showDOI       #1{#1}\fi
\ifx \showISBNx    \undefined \def \showISBNx     #1{\unskip}     \fi
\ifx \showISBNxiii \undefined \def \showISBNxiii  #1{\unskip}     \fi
\ifx \showISSN     \undefined \def \showISSN      #1{\unskip}     \fi
\ifx \showLCCN     \undefined \def \showLCCN      #1{\unskip}     \fi
\ifx \shownote     \undefined \def \shownote      #1{#1}          \fi
\ifx \showarticletitle \undefined \def \showarticletitle #1{#1}   \fi
\ifx \showURL      \undefined \def \showURL       {\relax}        \fi
\providecommand\bibfield[2]{#2}
\providecommand\bibinfo[2]{#2}
\providecommand\natexlab[1]{#1}
\providecommand\showeprint[2][]{arXiv:#2}

\bibitem[\protect\citeauthoryear{Abdul-Massih, Yoo, and Benes}{Abdul-Massih
  et~al\mbox{.}}{2017}]%
        {abdulmassih2017motionSR}
\bibfield{author}{\bibinfo{person}{Michel Abdul-Massih},
  \bibinfo{person}{Innfarn Yoo}, {and} \bibinfo{person}{Bedrich Benes}.}
  \bibinfo{year}{2017}\natexlab{}.
\newblock \showarticletitle{Motion Style Retargeting to Characters With
  Different Morphologies}.
\newblock \bibinfo{journal}{\emph{Comp.~Graph.~Forum}}  \bibinfo{volume}{36}
  (\bibinfo{year}{2017}), \bibinfo{pages}{86--99}.
\newblock


\bibitem[\protect\citeauthoryear{Aberman, Wu, Lischinski, Chen, and
  Cohen-Or}{Aberman et~al\mbox{.}}{2019}]%
        {aberman2019learning}
\bibfield{author}{\bibinfo{person}{Kfir Aberman}, \bibinfo{person}{Rundi Wu},
  \bibinfo{person}{Dani Lischinski}, \bibinfo{person}{Baoquan Chen}, {and}
  \bibinfo{person}{Daniel Cohen-Or}.} \bibinfo{year}{2019}\natexlab{}.
\newblock \showarticletitle{Learning Character-Agnostic Motion for Motion
  Retargeting in {2D}}.
\newblock \bibinfo{journal}{\emph{ACM Trans.~Graph.}} \bibinfo{volume}{38},
  \bibinfo{number}{4} (\bibinfo{year}{2019}), \bibinfo{pages}{75}.
\newblock


\bibitem[\protect\citeauthoryear{{Adobe Systems Inc.}}{{Adobe Systems
  Inc.}}{2018}]%
        {mixamo}
\bibfield{author}{\bibinfo{person}{{Adobe Systems Inc.}}}
  \bibinfo{year}{2018}\natexlab{}.
\newblock \bibinfo{title}{Mixamo}.
\newblock \bibinfo{howpublished}{https://www.mixamo.com}.
\newblock
\urldef\tempurl%
\url{https://www.mixamo.com}
\showURL{%
\tempurl}
\newblock
\shownote{Accessed: 2018-12-27.}


\bibitem[\protect\citeauthoryear{Baran, Vlasic, Grinspun, and
  Popovi{\'c}}{Baran et~al\mbox{.}}{2009}]%
        {baran2009semanticDT}
\bibfield{author}{\bibinfo{person}{Ilya Baran}, \bibinfo{person}{Daniel
  Vlasic}, \bibinfo{person}{Eitan Grinspun}, {and} \bibinfo{person}{Jovan
  Popovi{\'c}}.} \bibinfo{year}{2009}\natexlab{}.
\newblock \showarticletitle{Semantic Deformation Transfer}.
\newblock \bibinfo{journal}{\emph{ACM Trans. Graph.}} \bibinfo{volume}{28},
  \bibinfo{number}{3}, Article \bibinfo{articleno}{36} (\bibinfo{date}{July}
  \bibinfo{year}{2009}), \bibinfo{numpages}{6}~pages.
\newblock


\bibitem[\protect\citeauthoryear{Bruna, Zaremba, Szlam, and LeCun}{Bruna
  et~al\mbox{.}}{2013}]%
        {bruna2013spectral}
\bibfield{author}{\bibinfo{person}{Joan Bruna}, \bibinfo{person}{Wojciech
  Zaremba}, \bibinfo{person}{Arthur Szlam}, {and} \bibinfo{person}{Yann
  LeCun}.} \bibinfo{year}{2013}\natexlab{}.
\newblock \bibinfo{title}{Spectral Networks and Locally Connected Networks on
  Graphs}.
\newblock
\newblock
\showeprint[arxiv]{cs.LG/1312.6203}


\bibitem[\protect\citeauthoryear{Celikcan, Yaz, and {\c C}apin}{Celikcan
  et~al\mbox{.}}{2015}]%
        {celikcan2015exampleBasedRO}
\bibfield{author}{\bibinfo{person}{Ufuk Celikcan}, \bibinfo{person}{Ilker~O.
  Yaz}, {and} \bibinfo{person}{Tolga~K. {\c C}apin}.}
  \bibinfo{year}{2015}\natexlab{}.
\newblock \showarticletitle{Example-Based Retargeting of Human Motion to
  Arbitrary Mesh Models}.
\newblock \bibinfo{journal}{\emph{Comp.~Graph.~Forum}}  \bibinfo{volume}{34}
  (\bibinfo{year}{2015}), \bibinfo{pages}{216--227}.
\newblock


\bibitem[\protect\citeauthoryear{Choi and Ko}{Choi and Ko}{2000}]%
        {choi2000online}
\bibfield{author}{\bibinfo{person}{Kwang-Jin Choi} {and}
  \bibinfo{person}{Hyeong-Seok Ko}.} \bibinfo{year}{2000}\natexlab{}.
\newblock \showarticletitle{Online motion retargetting}.
\newblock \bibinfo{journal}{\emph{The Journal of Visualization and Computer
  Animation}} \bibinfo{volume}{11}, \bibinfo{number}{5} (\bibinfo{year}{2000}),
  \bibinfo{pages}{223--235}.
\newblock


\bibitem[\protect\citeauthoryear{Ci, Wang, Ma, and Wang}{Ci
  et~al\mbox{.}}{2019}]%
        {ci2019optimizingNS}
\bibfield{author}{\bibinfo{person}{Hai Ci}, \bibinfo{person}{Chunyu Wang},
  \bibinfo{person}{Xiaoxuan Ma}, {and} \bibinfo{person}{Yizhou Wang}.}
  \bibinfo{year}{2019}\natexlab{}.
\newblock \showarticletitle{Optimizing Network Structure for 3D Human Pose
  Estimation}. In \bibinfo{booktitle}{\emph{Proc.~ICCV 2019}}.
\newblock


\bibitem[\protect\citeauthoryear{Delhaisse, Esteban, Rozo, and
  Caldwell}{Delhaisse et~al\mbox{.}}{2017}]%
        {delhaisse2017transfer}
\bibfield{author}{\bibinfo{person}{Brian Delhaisse}, \bibinfo{person}{Domingo
  Esteban}, \bibinfo{person}{Leonel~Dario Rozo}, {and}
  \bibinfo{person}{Darwin~G. Caldwell}.} \bibinfo{year}{2017}\natexlab{}.
\newblock \showarticletitle{Transfer learning of shared latent spaces between
  robots with similar kinematic structure}. In
  \bibinfo{booktitle}{\emph{Proc.~IJCNN 2017}}. \bibinfo{pages}{4142--4149}.
\newblock


\bibitem[\protect\citeauthoryear{Feng, Huang, Xu, and Shapiro}{Feng
  et~al\mbox{.}}{2012}]%
        {feng2012automating}
\bibfield{author}{\bibinfo{person}{Andrew Feng}, \bibinfo{person}{Yazhou
  Huang}, \bibinfo{person}{Yuyu Xu}, {and} \bibinfo{person}{Ari Shapiro}.}
  \bibinfo{year}{2012}\natexlab{}.
\newblock \showarticletitle{Automating the transfer of a generic set of
  behaviors onto a virtual character}. In
  \bibinfo{booktitle}{\emph{International Conference on Motion in Games}}.
  Springer, \bibinfo{pages}{134--145}.
\newblock


\bibitem[\protect\citeauthoryear{Gao, Yang, Qiao, Lai, Rosin, Xu, and Xia}{Gao
  et~al\mbox{.}}{2018}]%
        {gao2018automaticUS}
\bibfield{author}{\bibinfo{person}{Lin Gao}, \bibinfo{person}{Jie Yang},
  \bibinfo{person}{Yi-Ling Qiao}, \bibinfo{person}{Yu-Kun Lai},
  \bibinfo{person}{Paul~L. Rosin}, \bibinfo{person}{Weiwei Xu}, {and}
  \bibinfo{person}{Shihong Xia}.} \bibinfo{year}{2018}\natexlab{}.
\newblock \showarticletitle{Automatic unpaired shape deformation transfer}.
\newblock \bibinfo{journal}{\emph{ACM Trans. Graph.}}  \bibinfo{volume}{37}
  (\bibinfo{year}{2018}), \bibinfo{pages}{237:1--237:15}.
\newblock


\bibitem[\protect\citeauthoryear{Gleicher}{Gleicher}{1998}]%
        {gleicher1998retargetting}
\bibfield{author}{\bibinfo{person}{Michael Gleicher}.}
  \bibinfo{year}{1998}\natexlab{}.
\newblock \showarticletitle{Retargetting motion to new characters}. In
  \bibinfo{booktitle}{\emph{Proc.~25th annual conference on computer graphics
  and interactive techniques}}. ACM, \bibinfo{pages}{33--42}.
\newblock


\bibitem[\protect\citeauthoryear{Gonzalez-Garcia, van~de Weijer, and
  Bengio}{Gonzalez-Garcia et~al\mbox{.}}{2018}]%
        {gonzalez2018image}
\bibfield{author}{\bibinfo{person}{Abel Gonzalez-Garcia},
  \bibinfo{person}{Joost van~de Weijer}, {and} \bibinfo{person}{Yoshua
  Bengio}.} \bibinfo{year}{2018}\natexlab{}.
\newblock \showarticletitle{Image-to-image translation for cross-domain
  disentanglement}. In \bibinfo{booktitle}{\emph{Advances in Neural Information
  Processing Systems}}. \bibinfo{pages}{1287--1298}.
\newblock


\bibitem[\protect\citeauthoryear{Hanocka, Hertz, Fish, Giryes, Fleishman, and
  Cohen-Or}{Hanocka et~al\mbox{.}}{2019}]%
        {hanocka2019meshcnn}
\bibfield{author}{\bibinfo{person}{Rana Hanocka}, \bibinfo{person}{Amir Hertz},
  \bibinfo{person}{Noa Fish}, \bibinfo{person}{Raja Giryes},
  \bibinfo{person}{Shachar Fleishman}, {and} \bibinfo{person}{Daniel
  Cohen-Or}.} \bibinfo{year}{2019}\natexlab{}.
\newblock \showarticletitle{{MeshCNN:} a network with an edge}.
\newblock \bibinfo{journal}{\emph{ACM Trans.~Graph.}} \bibinfo{volume}{38},
  \bibinfo{number}{4} (\bibinfo{year}{2019}), \bibinfo{pages}{1--12}.
\newblock


\bibitem[\protect\citeauthoryear{Holden, Saito, and Komura}{Holden
  et~al\mbox{.}}{2016}]%
        {holden2016deep}
\bibfield{author}{\bibinfo{person}{Daniel Holden}, \bibinfo{person}{Jun Saito},
  {and} \bibinfo{person}{Taku Komura}.} \bibinfo{year}{2016}\natexlab{}.
\newblock \showarticletitle{A deep learning framework for character motion
  synthesis and editing}.
\newblock \bibinfo{journal}{\emph{ACM Trans.~Graph.}} \bibinfo{volume}{35},
  \bibinfo{number}{4} (\bibinfo{date}{July} \bibinfo{year}{2016}),
  \bibinfo{pages}{138:1--11}.
\newblock


\bibitem[\protect\citeauthoryear{Holden, Saito, Komura, and Joyce}{Holden
  et~al\mbox{.}}{2015}]%
        {holden2015learning}
\bibfield{author}{\bibinfo{person}{Daniel Holden}, \bibinfo{person}{Jun Saito},
  \bibinfo{person}{Taku Komura}, {and} \bibinfo{person}{Thomas Joyce}.}
  \bibinfo{year}{2015}\natexlab{}.
\newblock \showarticletitle{Learning motion manifolds with convolutional
  autoencoders}. In \bibinfo{booktitle}{\emph{SIGGRAPH Asia 2015 Technical
  Briefs}}. ACM, \bibinfo{pages}{18:1--18:4}.
\newblock


\bibitem[\protect\citeauthoryear{Huang, Liu, Belongie, and Kautz}{Huang
  et~al\mbox{.}}{2018}]%
        {huang2018multimodal}
\bibfield{author}{\bibinfo{person}{Xun Huang}, \bibinfo{person}{Ming-Yu Liu},
  \bibinfo{person}{Serge Belongie}, {and} \bibinfo{person}{Jan Kautz}.}
  \bibinfo{year}{2018}\natexlab{}.
\newblock \showarticletitle{Multimodal unsupervised image-to-image
  translation}. In \bibinfo{booktitle}{\emph{Proc.~ECCV 2018}}.
  \bibinfo{pages}{172--189}.
\newblock


\bibitem[\protect\citeauthoryear{Jain, Zamir, Savarese, and Saxena}{Jain
  et~al\mbox{.}}{2015}]%
        {jain2015structuralRNNDL}
\bibfield{author}{\bibinfo{person}{Ashesh Jain}, \bibinfo{person}{Amir~Roshan
  Zamir}, \bibinfo{person}{Silvio Savarese}, {and} \bibinfo{person}{Ashutosh
  Saxena}.} \bibinfo{year}{2015}\natexlab{}.
\newblock \showarticletitle{{Structural-RNN}: Deep Learning on Spatio-Temporal
  Graphs}. In \bibinfo{booktitle}{\emph{Proc.~IEEE CVPR 2015}}.
  \bibinfo{pages}{5308--5317}.
\newblock


\bibitem[\protect\citeauthoryear{Jang, Kwon, Yu, Kim, and Kim}{Jang
  et~al\mbox{.}}{2018}]%
        {jang2018variational}
\bibfield{author}{\bibinfo{person}{Hanyoung Jang}, \bibinfo{person}{Byungjun
  Kwon}, \bibinfo{person}{Moonwon Yu}, \bibinfo{person}{Seong~Uk Kim}, {and}
  \bibinfo{person}{Jongmin Kim}.} \bibinfo{year}{2018}\natexlab{}.
\newblock \showarticletitle{A variational U-Net for motion retargeting}. In
  \bibinfo{booktitle}{\emph{SIGGRAPH Asia 2018 Posters}}.
\newblock


\bibitem[\protect\citeauthoryear{Kingma and Ba}{Kingma and Ba}{2014}]%
        {kingma2014adam}
\bibfield{author}{\bibinfo{person}{Diederik~P Kingma} {and}
  \bibinfo{person}{Jimmy Ba}.} \bibinfo{year}{2014}\natexlab{}.
\newblock \showarticletitle{Adam: A method for stochastic optimization}.
\newblock \bibinfo{journal}{\emph{arXiv preprint arXiv:1412.6980}}
  (\bibinfo{year}{2014}).
\newblock


\bibitem[\protect\citeauthoryear{Lee and Shin}{Lee and Shin}{1999}]%
        {lee1999hierarchical}
\bibfield{author}{\bibinfo{person}{Jehee Lee} {and} \bibinfo{person}{Sung~Yong
  Shin}.} \bibinfo{year}{1999}\natexlab{}.
\newblock \showarticletitle{A hierarchical approach to interactive motion
  editing for human-like figures}. In \bibinfo{booktitle}{\emph{Proc.~26th
  annual conference on computer graphics and interactive techniques}}. ACM
  Press/Addison-Wesley Publishing Co., \bibinfo{pages}{39--48}.
\newblock


\bibitem[\protect\citeauthoryear{Lim, Chang, and Choi}{Lim
  et~al\mbox{.}}{2019}]%
        {lim2019PMnetLO}
\bibfield{author}{\bibinfo{person}{Jongin Lim}, \bibinfo{person}{Hyung~Jin
  Chang}, {and} \bibinfo{person}{Jin~Young Choi}.}
  \bibinfo{year}{2019}\natexlab{}.
\newblock \showarticletitle{{PMnet:} learning of disentangled pose and movement
  for unsupervised motion retargeting}. In
  \bibinfo{booktitle}{\emph{Proc.~BMVC}}.
\newblock


\bibitem[\protect\citeauthoryear{Niepert, Ahmed, and Kutzkov}{Niepert
  et~al\mbox{.}}{2016}]%
        {Niepert:2016}
\bibfield{author}{\bibinfo{person}{Mathias Niepert}, \bibinfo{person}{Mohamed
  Ahmed}, {and} \bibinfo{person}{Konstantin Kutzkov}.}
  \bibinfo{year}{2016}\natexlab{}.
\newblock \showarticletitle{Learning Convolutional Neural Networks for Graphs}.
  In \bibinfo{booktitle}{\emph{Proc.~ICML 2016}}, Vol.~\bibinfo{volume}{48}.
  \bibinfo{pages}{2014--2023}.
\newblock


\bibitem[\protect\citeauthoryear{Pavllo, Feichtenhofer, Auli, and
  Grangier}{Pavllo et~al\mbox{.}}{2019}]%
        {pavllo2019modelingHM}
\bibfield{author}{\bibinfo{person}{Dario Pavllo}, \bibinfo{person}{Christoph
  Feichtenhofer}, \bibinfo{person}{Michael Auli}, {and} \bibinfo{person}{David
  Grangier}.} \bibinfo{year}{2019}\natexlab{}.
\newblock \showarticletitle{Modeling Human Motion with Quaternion-Based Neural
  Networks}.
\newblock \bibinfo{journal}{\emph{International Journal of Computer Vision}}
  (\bibinfo{date}{October} \bibinfo{year}{2019}), \bibinfo{pages}{1--18}.
\newblock


\bibitem[\protect\citeauthoryear{Seol, O'Sullivan, and Lee}{Seol
  et~al\mbox{.}}{2013}]%
        {seol2013creatureFO}
\bibfield{author}{\bibinfo{person}{Yeongho Seol}, \bibinfo{person}{Carol
  O'Sullivan}, {and} \bibinfo{person}{Jehee Lee}.}
  \bibinfo{year}{2013}\natexlab{}.
\newblock \showarticletitle{Creature features: online motion puppetry for
  non-human characters}. In \bibinfo{booktitle}{\emph{Proc.~SCA '13}}.
\newblock


\bibitem[\protect\citeauthoryear{Sumner and Popovi{\'c}}{Sumner and
  Popovi{\'c}}{2004}]%
        {sumner2004deformation}
\bibfield{author}{\bibinfo{person}{Robert~W. Sumner} {and}
  \bibinfo{person}{Jovan Popovi{\'c}}.} \bibinfo{year}{2004}\natexlab{}.
\newblock \showarticletitle{Deformation Transfer for Triangle Meshes}.
\newblock \bibinfo{journal}{\emph{ACM Trans. Graph.}} \bibinfo{volume}{23},
  \bibinfo{number}{3} (\bibinfo{date}{Aug.} \bibinfo{year}{2004}),
  \bibinfo{pages}{399--405}.
\newblock


\bibitem[\protect\citeauthoryear{Tak and Ko}{Tak and Ko}{2005}]%
        {tak2005physically}
\bibfield{author}{\bibinfo{person}{Seyoon Tak} {and}
  \bibinfo{person}{Hyeong-Seok Ko}.} \bibinfo{year}{2005}\natexlab{}.
\newblock \showarticletitle{A physically-based motion retargeting filter}.
\newblock \bibinfo{journal}{\emph{ACM Trans.~Graph.}} \bibinfo{volume}{24},
  \bibinfo{number}{1} (\bibinfo{year}{2005}), \bibinfo{pages}{98--117}.
\newblock


\bibitem[\protect\citeauthoryear{Villegas, Yang, Ceylan, and Lee}{Villegas
  et~al\mbox{.}}{2018}]%
        {villegas2018neural}
\bibfield{author}{\bibinfo{person}{Ruben Villegas}, \bibinfo{person}{Jimei
  Yang}, \bibinfo{person}{Duygu Ceylan}, {and} \bibinfo{person}{Honglak Lee}.}
  \bibinfo{year}{2018}\natexlab{}.
\newblock \showarticletitle{Neural Kinematic Networks for Unsupervised Motion
  Retargetting}. In \bibinfo{booktitle}{\emph{Proc.~IEEE CVPR}}.
  \bibinfo{pages}{8639--8648}.
\newblock


\bibitem[\protect\citeauthoryear{Wang, Ho, Shum, and Zhu}{Wang
  et~al\mbox{.}}{2019}]%
        {wang2019spatiotemporal}
\bibfield{author}{\bibinfo{person}{He Wang}, \bibinfo{person}{Edmond S.~L. Ho},
  \bibinfo{person}{Hubert P.~H. Shum}, {and} \bibinfo{person}{Zhanxing Zhu}.}
  \bibinfo{year}{2019}\natexlab{}.
\newblock \bibinfo{title}{Spatio-temporal Manifold Learning for Human Motions
  via Long-horizon Modeling}.
\newblock
\newblock
\showeprint[arxiv]{cs.GR/1908.07214}


\bibitem[\protect\citeauthoryear{Yamane, Ariki, and Hodgins}{Yamane
  et~al\mbox{.}}{2010}]%
        {yamane2010animatingNC}
\bibfield{author}{\bibinfo{person}{Katsu Yamane}, \bibinfo{person}{Yuka Ariki},
  {and} \bibinfo{person}{Jessica~K. Hodgins}.} \bibinfo{year}{2010}\natexlab{}.
\newblock \showarticletitle{Animating non-humanoid characters with human motion
  data}. In \bibinfo{booktitle}{\emph{Proc.~SCA '10}}.
\newblock


\bibitem[\protect\citeauthoryear{Yan, Xiong, and Lin}{Yan
  et~al\mbox{.}}{2018}]%
        {yan2018spatial}
\bibfield{author}{\bibinfo{person}{Sijie Yan}, \bibinfo{person}{Yuanjun Xiong},
  {and} \bibinfo{person}{Dahua Lin}.} \bibinfo{year}{2018}\natexlab{}.
\newblock \showarticletitle{Spatial temporal graph convolutional networks for
  skeleton-based action recognition}. In
  \bibinfo{booktitle}{\emph{Thirty-Second AAAI Conference on Artificial
  Intelligence}}.
\newblock


\bibitem[\protect\citeauthoryear{Zhu, Park, Isola, and Efros}{Zhu
  et~al\mbox{.}}{2017}]%
        {zhu2017unpaired}
\bibfield{author}{\bibinfo{person}{Jun-Yan Zhu}, \bibinfo{person}{Taesung
  Park}, \bibinfo{person}{Phillip Isola}, {and} \bibinfo{person}{Alexei~A
  Efros}.} \bibinfo{year}{2017}\natexlab{}.
\newblock \showarticletitle{Unpaired image-to-image translation using
  cycle-consistent adversarial networks}. In
  \bibinfo{booktitle}{\emph{Proc.~ICCV 2017}}. \bibinfo{pages}{2223--2232}.
\newblock


\end{thebibliography}

\appendix

\section{Network Architectures}

\label{appendix}

In this section we describe the details for the network architectures. 

Table~\ref{tab:aware_arch} describes the architecture for our skeleton-aware network,
where \texttt{Conv, LReLU, AP, UP} and  \texttt{UpS} denote convolution, leaky ReLU, average skeletal pooling, skeletal unpooling and temporal linear upsampling layers, respectively. 
All of the convolution layers use reflected padding. $k$ is the kernel width, $s$ is the stride, and the number of input and output channels per joint, and number of input and output joint number is reported in the rightmost column.

\begin{table}
\begin{center}
    \begin{tabular}{ l  l  c  c  c  c  }
    \toprule
        Name & Layers & $k$ & $s$ & i/o_c & i/o_j\\
    \toprule 
        $E_A^Q$  & \texttt{Conv + LReLU + AP}  & $15$ & 2 & $4/8$ & $28/18$\\
         & \texttt{Conv + LReLU + AP} & $15$ & 2 & $8/16$ & $18/7$ \\
    \midrule 
        $D_A^Q$  & \texttt{UP + UpS + Conv + LReLU} & $15$ & 1 & $16/8$ & $7/18$\\
         &\texttt{UP + UpS + Conv} & $15$ & 1 & $8/4$ & $18/28$\\
    \midrule 
       $E_A^S$  & \texttt{Conv + LReLU + AP} & $1$ & 1 & $3/8$ & $28/18$ \\
    \midrule 
        $E_B^Q$  & \texttt{Conv + LReLU + AP}  & $15$ & 2 & $4/8$ & $22/12$\\
        & \texttt{Conv + LReLU + AP} & $15$ & 2 & $8/16$ & $12/7$ \\
    \midrule 
        $D_B^Q$  & \texttt{UP + UpS + Conv + LReLU} & $15$ & 1 & $16/8$ & $7/12$\\
        &\texttt{UP + UpS + Conv} & $15$ & 1 & $8/4$ & $12/22$\\
    \midrule 
        $E_B^S$  & \texttt{Conv + LReLU + AP} & $1$ & 1 & $3/8$ & $22/12$ \\
    \bottomrule
    \end{tabular}
\end{center}
\caption{Architecture for skeleton-aware network}
\label{tab:aware_arch}
\end{table}

\begin{table}
\begin{center}
    \begin{tabular}{ l  l  c  c  c  }
    \toprule
        Name & Layers & $k$ & $s$ & i/o_c\\
    \toprule 
        $E_A^Q$  & \texttt{Conv + LReLU}  & $15$ & 2 & $112/144$\\
         & \texttt{Conv + LReLU} & $15$ & 2 & $144/112$ \\
    \midrule 
        $D_A^Q$  & \texttt{UpS + Conv + LReLU} & $15$ & 1 & $112/144$\\
         &\texttt{UpS + Conv} & $15$ & 1 & $144/112$\\
    \midrule 
       $E_A^S$  & \texttt{Conv + LReLU} & $1$ & 1 & $84/144$ \\
    \midrule 
        $E_B^Q$  & \texttt{Conv + LReLU }  & $15$ & 2 & $88/96$\\
        & \texttt{Conv + LReLU} & $15$ & 2 & $96/122$\\
    \midrule 
        $D_B^Q$  & \texttt{UpS + Conv + LReLU} & $15$ & 1 & $122/96$\\
        &\texttt{UpS + Conv} & $15$ & 1 & $96/88$ \\
    \midrule 
        $E_B^S$  & \texttt{Conv + LReLU} & $1$ & 1 & $66/96$\\
    \bottomrule
    \end{tabular}
\end{center}
\caption{Architecture for regular skeleton-unaware network}
\label{tab:unaware_arch}
\end{table}

Table~\ref{tab:unaware_arch} describes the architecture of the regular skeleton-unaware network, where \texttt{Conv}, \texttt{LReLU}, and \texttt{UpS} denote 1-D temporal convolution, LeakyReLU, and temporal linear upsampling layers, respectively. It is implemented by achieving  the same total number of channels (i.e., $\mbox{number of joints} \times \mbox{channels per joint}$) as in the skeleton-aware network.

Discriminators $C_A, C_B$ are implemented as patch-GAN discriminators. They share the same architecture as of $E_A, E_B$, except using sigmoid as the last activation.

\end{document}